\documentclass[10pt,twocolumn,letterpaper]{article}

\usepackage{iccv}
\usepackage{times}
\usepackage{epsfig}
\usepackage{graphicx}
\usepackage{amsmath}
\usepackage{amssymb}

\usepackage[pagebackref=true,breaklinks=true,letterpaper=true,colorlinks,bookmarks=false]{hyperref}

\iccvfinalcopy 

\def\iccvPaperID{13} 

\ificcvfinal\fi

\makeatletter
\newcommand{\printfnsymbol}[1]{%
  \textsuperscript{\@fnsymbol{#1}}%
}

\begin{document}



\title{UniNet: A Unified Scene Understanding Network and Exploring Multi-Task Relationships through the Lens of Adversarial Attacks}

\author{Naresh Kumar Gurulingan\thanks{Equal contribution.}, Elahe Arani\footnotemark[1], Bahram Zonooz \\
Advanced Research Lab, NavInfo Europe, Eindhoven, The Netherlands \\
{\tt\small \{naresh.gurulingan, elahe.arani\}@navinfo.eu}, {\tt\small bahram.zonooz@gmail.com} \\
}

\maketitle
\ificcvfinal\thispagestyle{empty}\fi

\begin{abstract}
Scene understanding is crucial for autonomous systems which intend to operate in the real world. Single task vision networks extract information only based on some aspects of the scene. In multi-task learning (MTL), on the other hand, these single tasks are jointly learned, thereby providing an opportunity for tasks to share information and obtain a more comprehensive understanding. To this end, we develop UniNet, a unified scene understanding network that accurately and efficiently infers vital vision tasks including object detection, semantic segmentation, instance segmentation, monocular depth estimation, and monocular instance depth prediction. As these tasks look at different semantic and geometric information, they can either complement or conflict with each other. Therefore, understanding inter-task relationships can provide useful cues to enable complementary information sharing. We evaluate the task relationships in UniNet through the lens of adversarial attacks based on the notion that they can exploit learned biases and task interactions in the neural network. Extensive experiments on the Cityscapes dataset, using untargeted and targeted attacks reveal that semantic tasks strongly interact amongst themselves, and the same holds for geometric tasks. Additionally, we show that the relationship between semantic and geometric tasks is asymmetric and their interaction becomes weaker as we move towards higher-level representations. Code has been made public\footnote{\url{github.com/NeurAI-Lab/UniNet}}.
\end{abstract}

\section{Introduction}
Scene understanding concerns the breakdown of a perceived scene into meaningful attributes which help autonomous systems understand and operate in the real world. With the aid of vision tasks which infer semantic and geometric information from monocular images, autonomous systems can deduce various relevant aspects of the world. While deep learning has provided numerous breakthroughs in single tasks such as object detection \cite{Girshick2015FastR,Redmon2017YOLO9000BF,Liu2016SSDSS,tian2019fcos}, semantic segmentation \cite{7298965,7803544,10.1007/978-3-030-01234-2_49}, instance segmentation \cite{He2017MaskR,Zhang2020MaskEF}, monocular depth estimation \cite{Alhashim2018,Yin2019EnforcingGC}, and monocular instance depth prediction \cite{Chen2018DrivingSP}, multi-task learning (MTL) and inter-task relationships have not been fully explored.




A multi-task network could be designed to have either a feature representation shared for all tasks \cite{Chen2018DrivingSP, Xu2018PADNetMG} or task-specific feature representations \cite{Misra2016CrossStitchNF}. While both designs benefit from complementary task information, the former provides added benefits such as reduced memory usage and inference time. We propose UniNet, a unified multi-task scene understanding network which is designed to jointly predict five crucial vision tasks - object detection (OD), semantic segmentation (SS), instance segmentation (IS), depth estimation (D), and instance depth prediction (ID). UniNet is based on an asymmetric encoder-decoder architecture, where the encoder is shared between all tasks while the decoder is only used for the lighter tasks. Hence, it provides an acceptable trade-off between accuracy and inference efficiency.

Learning multiple tasks together serves as an inductive bias prioritizing a learned representation that favours all tasks \cite{Caruana1993MultitaskLA}. However, the training procedure must factor in inter-task relationships to effectively search for this representation. Related tasks would promote complementary information sharing while unrelated tasks would interfere with each other, hindering attainment of an optimal shared representation. Therefore, understanding inter-task relationships can provide insights to leverage complementary task information. We consider UniNet as a simple and effective multi-task framework containing a wide variety of vision tasks to study the interaction between different tasks and their relatedness. 



Existing works define task relatedness using similarity between features learned by single-task networks \cite{Dwivedi2019RepresentationSA}, using empirical results obtained when tasks are jointly learned \cite{Standley2020WhichTS}, and based on domain knowledge such as the alignment of semantic edges and depth discontinuities \cite{v2020revisiting}. Adversarial attacks center around the efficacy of fooling neural networks by making imperceptible changes to the input image. We hypothesise that these attacks can be used to study task relationships as they exploit learned biases in the neural networks. We therefore study task relationships in a multi-task network through the lens of adversarial attacks.





Using adversarial attacks, we study multi-task relationships and uncover intriguing findings such as - (1) semantic tasks interact strongly amongst themselves and the same holds for geometric tasks, (2) semantic and geometric tasks do not affect each other equally, and (3) while the interaction between semantic and geometric tasks is strong in low-level representations, it becomes weaker in high-level representations. Additionally, we find that intra-task interaction between bounding box classification and regression in object detection promotes a bias towards object shapes. The contributions of this work are summarized as follows:
\begin{itemize}
    \item We propose UniNet, a unified multi-task scene understanding network designed to infer five tasks which shows competitive performance with existing multi-task approaches while being efficient at inference.
    \item We uncover inter-task relationships in multi-task networks (UniNet and a modified version of MTI-Net \cite{vandenhende2020mti}) with the aid of adversarial attacks such as Projected Gradient Descent (PGD) and semantic category hiding. Specifically, semantic tasks and geometric tasks show greater interaction in low-level representation relative to high-level representation.
    \item We show that intra-task relationship between classification and regression in object detection induces a desirable bias towards object shapes.
\end{itemize}

\section{Related Works}

\subsection{Single-Task Learning}

Existing object detection methods use two-stage \cite{Girshick2014RichFH,Girshick2015FastR,Ren2015FasterRT}, one-stage anchor-based \cite{Redmon2017YOLO9000BF,Liu2016SSDSS} or one-stage anchor-free \cite{tian2019fcos} approaches. We use one-stage anchor free FCOS \cite{tian2019fcos} for object detection. Long \etal \cite{7298965} introduced fully convolutional network (FCN) for semantic segmentation. Since then, a number of encoder-decoder based architectures \cite{7803544,10.1007/978-3-319-24574-4_28,Paszke2016ENetAD,10.1007/978-3-030-01234-2_49} have been proposed. Encoder-decoder based architectures have also been used for supervised depth estimation \cite{Alhashim2018,Yin2019EnforcingGC}. UniNet uses an encoder-decoder based architecture for both semantic segmentation and depth estimation. Instance segmentation approaches are generally object detection free \cite{wang2020solo} or object detection based. The object detection based approaches further differ into region based \cite{He2017MaskR, Lee2020CenterMaskRA} or encoding based \cite{Xu_2019_ICCV,Zhang2020MaskEF} approaches. We use MEInst \cite{Zhang2020MaskEF} approach which encodes instance masks using PCA. Chen \etal \cite{Chen2018DrivingSP} proposed the instance depth prediction task where they predict a single value denoting the median depth of a predicted instance.

\subsection{Multi-Task Learning and Task Relationships}

Multi-task learning concerns the joint prediction of multiple tasks. Existing works concentrate on directions such as designing the architecture \cite{Liu2019EndToEndML,Xu2018PADNetMG} and designing a loss balancing strategy \cite{Kendall2018MultitaskLU,Liu2019EndToEndML,Guo2018DynamicTP,Chennupati2019MultiNetMF}. Sharing a feature extractor between different task specific heads is a common architecture design approach. Likewise, tasks in UniNet share a feature extractor but segmentation and depth heads additionally share the decoder. Loss balancing is required to balance different loss scales and to encourage complementary information sharing. We use geometric loss strategy \cite{Chennupati2019MultiNetMF} for loss balancing. 

Standley \etal\cite{Standley2020WhichTS} looked at task relationships in a multi-task setting. They took an empirical approach and trained models with several combination of tasks. Tasks in combinations which result in low total loss are considered to have affinities with each other (are related). Other works consider task relationships for transfer learning \cite{taskonomy2018, Dwivedi2019RepresentationSA, Song2019DeepMT, Song_2020_CVPR}. Zamir \etal\cite{taskonomy2018} studied task transfer relationships using an empirical approach by transferring across different single task networks. Dwivedi \etal\cite{Dwivedi2019RepresentationSA} used similarity between learned features in single task networks using Representation Similarity Analysis (RSA) as task relatedness. Attribution maps have also been used to study transfer relationships on the basis that similar tasks look at similar input image regions \cite{Song2019DeepMT}\cite{Song_2020_CVPR}. Contrary to these works, we study task relationships using adversarial attacks.

\subsection{Adversarial Attacks}
Szegedy \etal\cite{42503} showed the existence of perturbations which can fool an image classifier while making no human perceivable changes to the input image. This process of fooling is called adversarial attacks. Fast Gradient Sign Method (FGSM) \cite{43405}, Iterative FGSM \cite{DBLP:conf/iclr/KurakinGB17}, Projected Gradient Descent (PGD) \cite{madry2018towards}, and others have explored stronger or varied attack types. 


Arnab \etal \cite{Arnab2018OnTR} studied robustness of semantic segmentation models through the use of residual connections, multi-scale processing, and transformations on adversarial images. Metzen \etal \cite{Metzen2017UniversalAP} created universal perturbations which force predicted segmentation of different images to resemble a static scene. They also showed that persons can be hidden from segmentation predictions. Likewise, Wong \etal \cite{NEURIPS2020_609e9d4b} hid instances from depth predictions. Xie \etal \cite{Xie2017AdversarialEF} proposed Dense Adversary Generation (DAG) intended to change the class of a set of predictions in object detection and semantic segmentation. 

Klingner \etal \cite{Klingner2020ImprovedNA} observed that the adversarial robustness of semantic segmentation improved when trained with self-supervised depth estimation. Mao \etal \cite{Mao2020MultitaskLS} posited that multi-task learning reduces the ease of attaining perturbations which attack all tasks, thereby providing inherent robustness. Contrary to these works, we study multi-task relationships instead of robustness.

\section{Unified Network (UniNet)}

\begin{figure}[t]
\centering
   \includegraphics[width=1\linewidth]{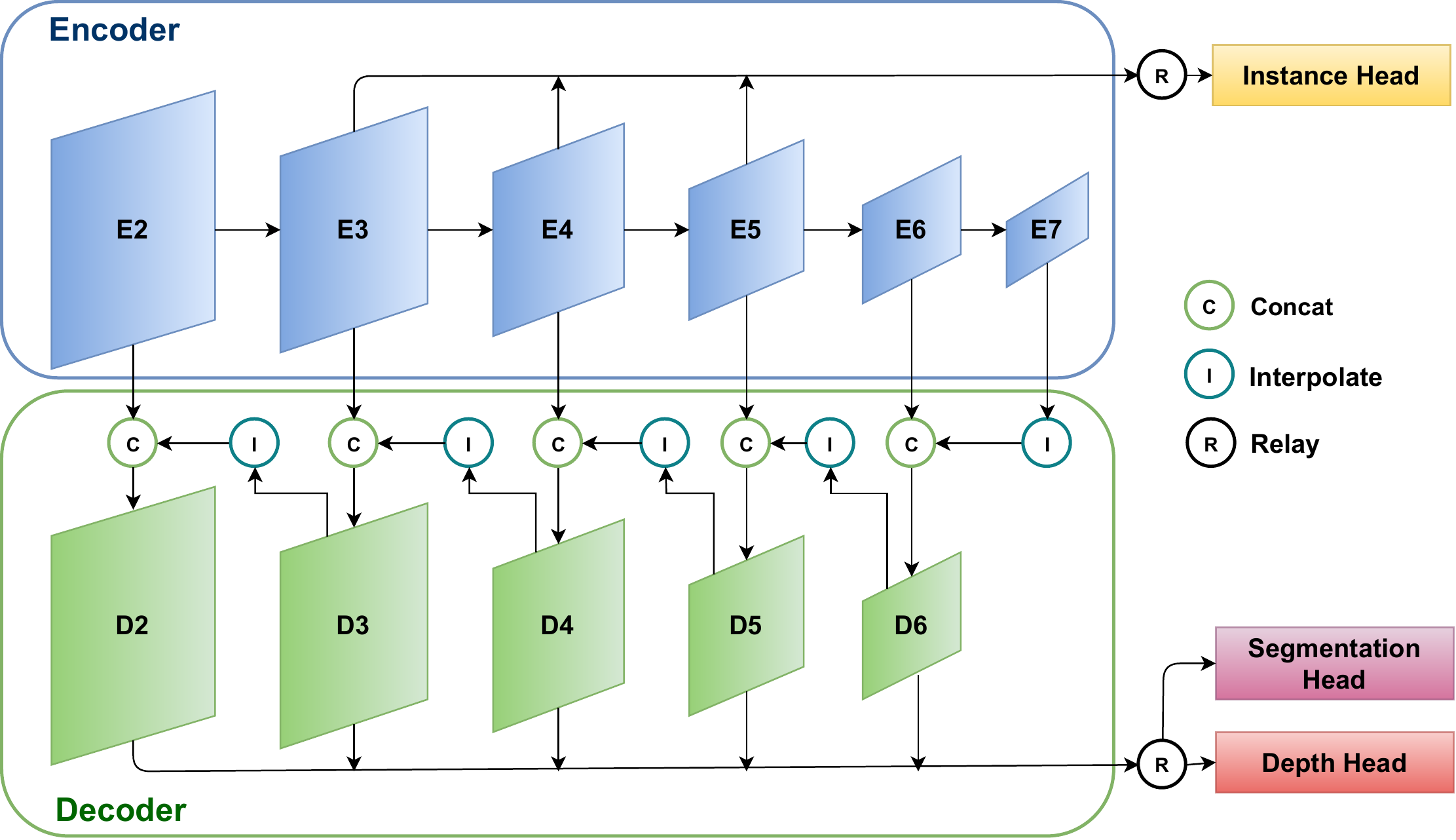}
   \caption{UniNet architecture: The encoder feature maps E3, E4, and E5 (blue) are relayed (R) to the instance head (yellow) which includes predictions of object detection, instance segmentation and instance depth. The decoder features D2 to D6 (green) are relayed to both the semantic segmentation (purple) and depth estimation head (red). Best viewed in color.}
\label{fig:uninet}
\end{figure}

UniNet is a multi-task network designed for scene understanding and infers tasks such as object detection, semantic segmentation, instance segmentation, depth estimation, and instance depth prediction. Object detection \cite{Girshick2015FastR, Redmon2017YOLO9000BF, Liu2016SSDSS, tian2019fcos} and instance segmentation \cite{He2017MaskR, Zhang2020MaskEF} detect and localize each occurrence of countable objects in the scene such as persons and cars. On the contrary, semantic segmentation \cite{7298965, 7803544, 10.1007/978-3-030-01234-2_49} is a per-pixel classification task which does not provide instance information. All three tasks provide abstract semantic information about the scene. In monocular depth estimation \cite{Alhashim2018, Yin2019EnforcingGC}, the depth of every input image pixel relative to the camera frame is predicted. Instance depth prediction \cite{Chen2018DrivingSP} involves predicting a single depth value for every countable object instance in the scene. Both depth estimation and instance depth prediction provide geometric information about the scene. Collectively, these tasks provide a comprehensive understanding of the scene.

\subsection{Architecture}
UniNet is based on asymmetric encoder-decoder architecture and is designed with a focus on efficient inference. In addition to the encoder and decoder, UniNet comprises of an instance head, a segmentation head and a depth head (Figure \ref{fig:uninet}). The decoder is shared between both the segmentation and the depth heads while the instance head branches from the encoder. The encoder features E2 to E7 are successively extracted from the input image. The decoder feature D6 is obtained by applying a residual block on the channel-wise concatenation of E6 and bi-linearly up-sampled E7. Decoder features D5, D4, D3 and D2 are also obtained in a similar fashion. E3, E4 and E5 are passed to the instance head while D2, D3, D4, D5 and D6 are passed to the semantic segmentation and depth estimation head.

\begin{figure*}
    \centering
        \begin{subfigure}{.55\textwidth}
          \centering
          \includegraphics[width=.95\linewidth]{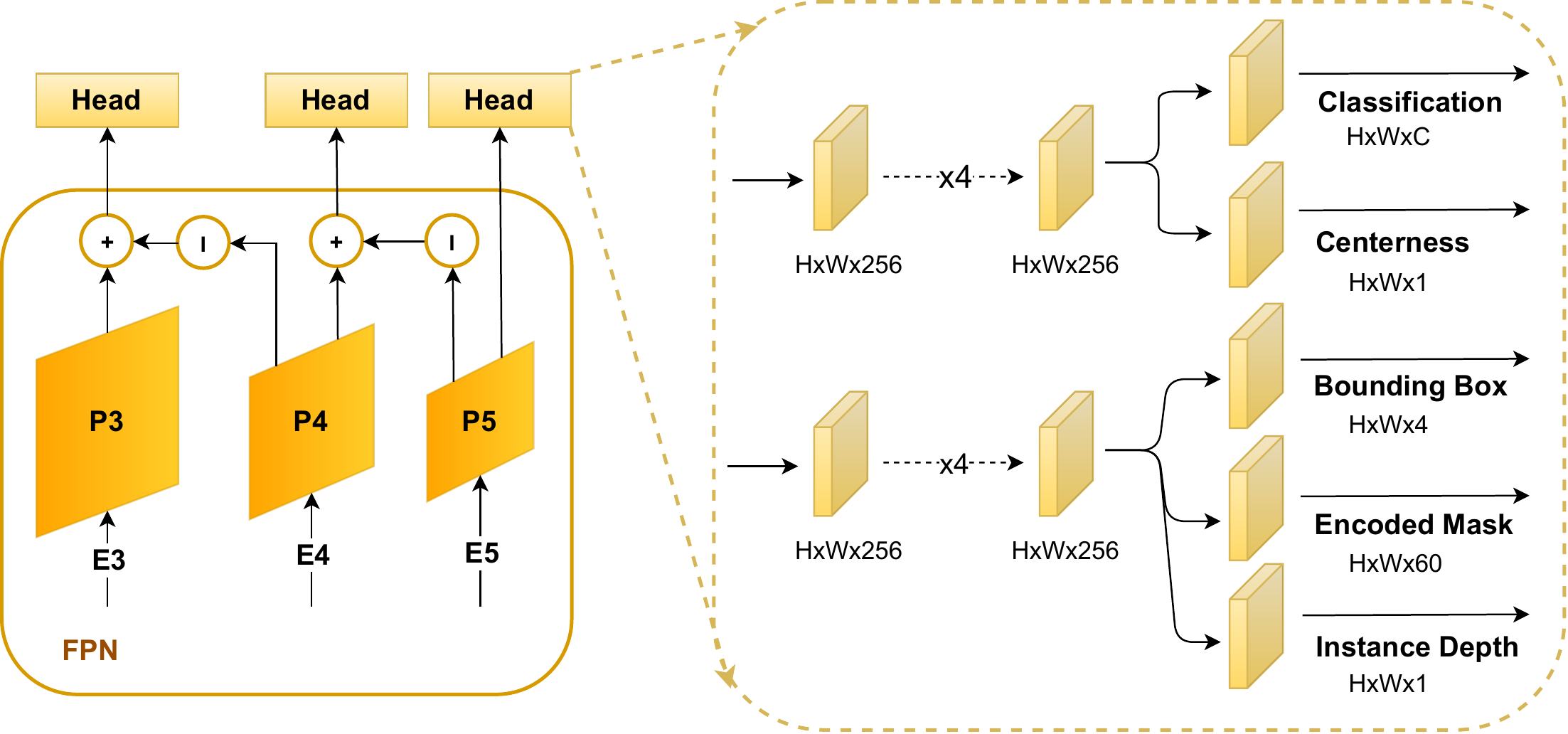}
          \caption{Instance Head}
          \label{fig:inst_head}
        \end{subfigure}%
        \begin{subfigure}{.22\textwidth}
          \centering
          \includegraphics[width=1\linewidth]{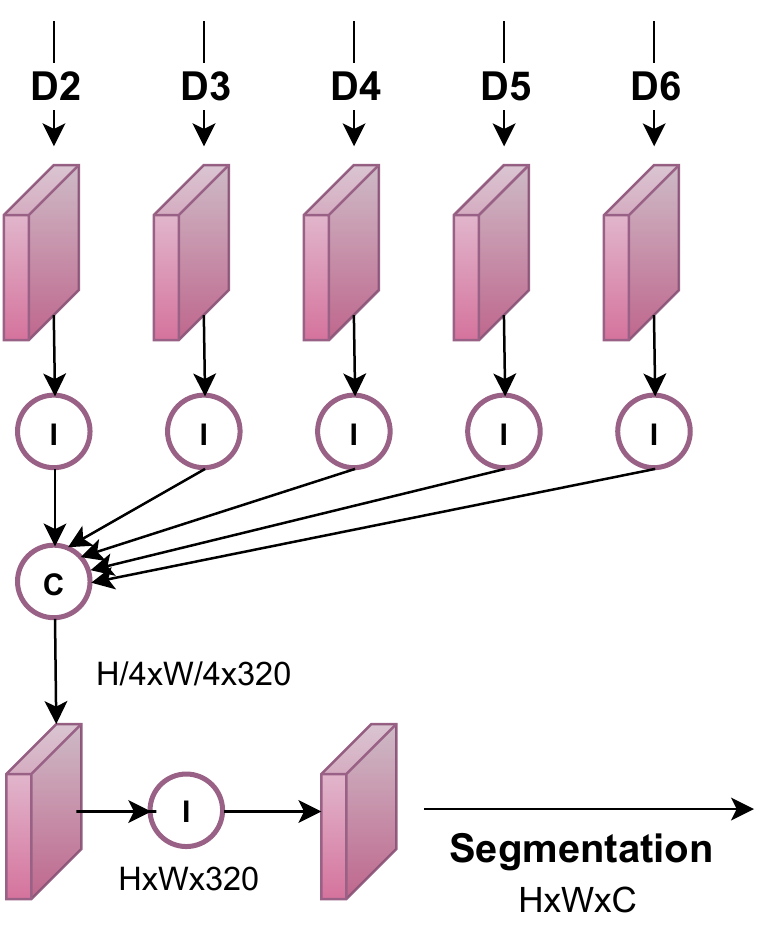}
          \caption{Segmentation Head}
          \label{fig:seg_head}
        \end{subfigure}
        \begin{subfigure}{.2\textwidth}
          \centering
          \includegraphics[width=1\linewidth]{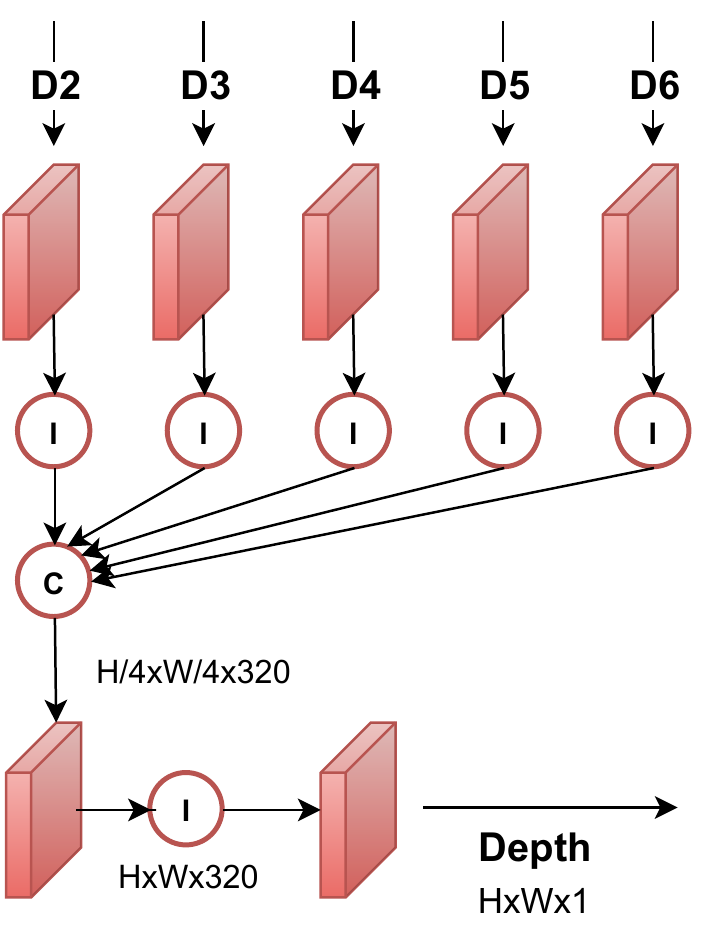}
          \caption{Depth Head}
          \label{fig:depth_head}
        \end{subfigure}
   \caption{Architecture of the heads. (a) depicts the FCOS \cite{tian2019fcos} based instance head. For each HxW feature map location, a classification logit, distance to the instance center, a bounding box, an encoded instance mask and a median instance depth is predicted. (b) and (c) depict the semantic segmentation and depth estimation head, respectively. \textbf{H, W} in (b) and (c) refer to the height and width of the original image. \textbf{C} refers to the number of classes.}
\label{fig:heads}
\end{figure*}

Figure \ref{fig:heads} shows the different heads in UniNet. The instance head (Figure \ref{fig:inst_head}) represents a collection of three instance tasks namely object detection, instance segmentation and instance depth prediction. A Feature Pyramid Network (FPN) \cite{Lin2017FeaturePN} is used to obtain features P3, P4 and P5 from E3, E4 and E5, respectively. Each location in P3, P4 and P5 serves as a point anchor, based on which classification logits, bounding box, and centerness values as in FCOS \cite{tian2019fcos}, encoded instance masks as in MEInst \cite{Zhang2020MaskEF} and median depth are predicted in a dense manner. In the semantic segmentation head (Figure \ref{fig:seg_head}), the channels of D2, D3, D4, D5 and D6 are first reduced to 64 and bi-linearly interpolated to $1/4^{th}$ of the image resolution. All the resultant feature maps are concatenated and provided as input to a convolution layer whose output is bi-linearly interpolated to image resolution. Finally, a convolution layer predicts the segmentation logits. The depth estimation head which predicts pixel-wise depth is architecturally similar to the semantic segmentation head, but the weights are not shared.

\subsection{Multi-Task Objective}

\textbf{Object detection}. Object detection is trained using box regression loss $\mathcal{L}_{reg}$, box classification loss $\mathcal{L}_{cls}$, and centerness loss $\mathcal{L}_{cent}$. We use the varifocal loss \cite{zhang2020varifocalnet} as $\mathcal{L}_{cls}$. Following FCOS \cite{tian2019fcos}, we use GIoU loss and binary cross entropy loss as $\mathcal{L}_{reg}$ and $\mathcal{L}_{cent}$, respectively. 

\textbf{Semantic segmentation}. We use a class balanced cross entropy loss ($\mathcal{L}_{seg}$) as the semantic segmentation loss.

\textbf{Instance segmentation}. We use the Mean Squared Error (MSE) between predictions and encoded ground truth masks ($\mathcal{L}_{is}$). The Principal Component Analysis (PCA) parameters obtained using training data is used to encode and decode instance masks following MEInst \cite{Zhang2020MaskEF}.

\textbf{Depth estimation}. The Root Mean Square Error (RMSE) between ground truth and prediction depth maps ($\mathcal{L}_{depth}$) is used as the the depth loss. 

\textbf{Instance depth prediction}. We use the $l_1$ loss between predicted median depth and ground truth median depth ($\mathcal{L}_{id}$).

\textbf{MTL Loss}. The MTL loss ($\mathcal{L}_{MTL}$) is the combination of all losses used to train UniNet. To balance the different losses involved, we first use fixed weights $\lambda_i$s to scale the losses and then use the geometric loss strategy \cite{Chennupati2019MultiNetMF}. In general, we observed that this balancing strategy provides an overall improvement in multi-task performance.
\begin{equation}
\mathcal{L}_{\text {MTL}}=\prod_{i} \sqrt[n]{\lambda_{i}\mathcal{L}_{i}}
\label{eq:mtlloss}
\end{equation}
\noindent
$n$ refers to the total number of losses required to be minimized and $i$ $\epsilon$ \{$reg$, $cls$, $cent$, $seg$, $is$, $depth$, $id$\}.

\section{Multi-Task relationships through Adversarial Attacks}

Adversarial attacks exploit vulnerabilities in learned neural network representations with the intent of forcing wrong predictions. In general, imperceptible perturbations obtained by optimizing an attack loss is added to the input image. For untargeted attacks, the same loss used to train the network is used as the attack loss. For targeted attack, the attack loss involves a target which has been modified in a certain manner to force a specific inference. For example, all ``persons" in the ground truth can be modified as ``road" forcing the network to predict ``persons" as ``road". 

Intuitively, these attacks leverage biases in the learned representation, resulting in perturbations which can fool the network. In multi-task networks, the architecture provides an inductive bias which enforces a feature space to be shared among tasks. This shared feature space facilitates interaction between the tasks which can be exploited by attacks. These intuitions present adversarial attacks as an intriguing front to study and evaluate task relationships in multi-task networks. We use the following attack methods.

\subsection{Projected Gradient Descent (PGD)}
\label{sec:pgd}

PGD \cite{madry2018towards} is an untargeted attack where the input image is updated to maximize a given loss. The multi-task loss used to train UniNet is composed of seven individual losses. We group these losses into semantic loss in Eq. \ref{eq:semanticloss} and geometric loss in Eq. \ref{eq:geometricloss}. Each of the individual losses, the multi-task loss (Eq. \ref{eq:mtlloss}), the semantic loss, and the geometric loss are all used as the PGD attack objective to study the multi-task relationships.
\begin{equation}
\mathcal{L}_{\text{semantic}}=\mathcal{L}_{reg} + \mathcal{L}_{cls} + \mathcal{L}_{seg} + \mathcal{L}_{is},
\label{eq:semanticloss}
\end{equation}
\begin{equation}
\mathcal{L}_{\text{geometric}}=\mathcal{L}_{depth} + \mathcal{L}_{id}.
\label{eq:geometricloss}
\end{equation}

\subsection{Dense Adversary Generation (DAG)} 
\label{sec:dagintro}

DAG \cite{Xie2017AdversarialEF} is an untargeted attack aimed at attaining perturbations which force the predictions of a target set to be different from the ground truth class. For object detection, we take the bounding boxes predicted in all locations of the detection head as the target set. For semantic segmentation, all the pixels are part of the target set. However, we modify DAG to perform targeted attacks. Specifically, we use the same algorithm as in \cite{Xie2017AdversarialEF} but modify the $\pi$ function as in Eq. \ref{eq:targeteddag}, where $c_1$ and $c_2$ are any two valid prediction classes. Essentially, the attack aims at swapping two selected prediction classes.
\begin{equation}
\pi(c_1) = c_2, \pi(c_2) = c_1.
\label{eq:targeteddag}
\end{equation}


\subsection{Semantic Category Hiding}
Metzen \etal \cite{Metzen2017UniversalAP} attack semantic segmentation with the objective of hiding a semantic category such as ``persons" from a segmentation prediction. All ``person" pixels are replaced with the nearest ``not person" pixels and is used as the target in the attack objective. The objective also includes a weight term which prioritizes retaining ``non person" pixels or hiding ``person" pixels. We use the same attack objective to hide persons from predicted segmentation maps without any weight term. Wong \etal \cite{NEURIPS2020_609e9d4b} hide instances from depth maps with the help of ground truth instance masks. However, since we jointly predict segmentation and depth, we hide pixels in the depth map which correspond to ``person" pixels in predicted segmentation map.

\section{Experiments}

\begin{table*}
\centering
\resizebox{\textwidth}{!}{%
\begin{tabular}{|l|cc|cccc|c|cc|cccc|c|}
\hline

 & \multicolumn{7}{|c|}{Cityscapes} & \multicolumn{7}{|c|}{NYUv2} \\ \cline{2-15}
Method & \makecell{SS \\ (mIoU)$\uparrow$} & \makecell{D \\ (RMSE)$\downarrow$} & \makecell{MAC \\ (G)} & \makecell{\#P \\ (M)} & \makecell{FPS \\} & \makecell{Energy \\ (J)} & \makecell{$\Delta_{MTL}$ \\ (\%)$\uparrow$} & \makecell{SS \\ (mIoU)$\uparrow$} & \makecell{D \\ (RMSE)$\downarrow$} & \makecell{MAC \\ (G)} & \makecell{\#P \\ (M)} & \makecell[t]{FPS} & \makecell{Energy \\ (J)} & \makecell{$\Delta_{MTL}$ \\ (\%)$\uparrow$} \\ \hline 

Single task \cite{v2020revisiting} & 64.96 & 5.802 & 35 & 20 & 73 & 2.6 & +0.00 & 40.59 & 50.27 & 20 & 20 & 121 & 2.0 & +0.00 \\ 
MTL-baseline \cite{v2020revisiting} & 65.14 & 5.890 & 37 & 25 & 67 & 2.8 & -0.62 & 40.33 & 48.51 & 22 & 25 & 109 & 2.2 & +1.43 \\ \hline
PADNet \cite{Xu2018PADNetMG} & 73.86 & 5.680 & 300 & 23 & 16 & 3.7 & +7.90 & 38.95 & 52.81 & 176 & 23 & 24 & 3.3 & -4.55 \\ 
PADNet$^\dagger$ \cite{Xu2018PADNetMG} & 73.47 & 5.630 & \textbf{137} & \textbf{18} & \textbf{27} & \textbf{3.1} & +8.03 & 38.03 & 53.90 & \textbf{80} & \textbf{18} & \textbf{45} & \textbf{2.9} & -6.76 \\
MTI-Net \cite{vandenhende2020mti} & \textbf{76.68} & \textbf{5.129} & 433 & 99 & 10 & 5.0 & \textbf{+14.82} & \textbf{43.27} & \textbf{45.43} & 254 & 99 & 13 & 4.8 & \textbf{+8.12} \\ 
MTI-Net$^\dagger$ \cite{vandenhende2020mti} & 75.90 & 5.163 & 235 & 53 & 19 & 3.9 & +3.93 & 42.50 & 46.63 & 137 & 53 & 23 & 3.9 & +5.97 \\ \hline
Cross-stitch \cite{Misra2016CrossStitchNF} & 65.31 & 5.743 & 69 & 40 & 36 & \textbf{3.4} & +0.78 & 40.09 & 48.49 & 41 & 40 & 59 & 4.0 & +1.15 \\ 
MTAN \cite{Liu2019EndToEndML} & 65.70 & 5.862 & 42 & 26 & 51 & 3.8 & +0.05 & 40.04 & 48.34 & 25 & 26 & 81 & \textbf{2.9} & +1.24 \\ 
UniNet & \textbf{74.49} & \textbf{5.379} & \textbf{37} & \textbf{18} & \textbf{48} & 3.5 & \textbf{+10.98} & \textbf{40.90} & \textbf{47.08} & \textbf{22} & \textbf{18} & \textbf{81} & \textbf{2.9} & \textbf{+3.55} \\ \hline

\end{tabular}}
\vspace*{1mm}
\caption{Comparing UniNet with state-of-the-art multi-task models on Cityscapes and NYUv2 datasets. $\Delta_{MTL}$ represents the multi-task performance metric proposed in \cite{v2020revisiting}. MTI-Net and PADNet use additional auxiliary tasks. We also include results for MTI-Net and PADNet without auxiliary tasks (represented by $\dagger$).}
\label{table:sota}
\end{table*}

\begin{table}
\centering
\resizebox{\columnwidth}{!}{%
\begin{tabular}{|c|c|c|c|c|c|}
\hline
Task & \makecell{Five \\ tasks} & \makecell{Single \\ task} & OD+IS & OD+ID & SS+D \\ \hline
\makecell{OD \\ ($\text{mAP}^b$)} & \textbf{38.93}{\scriptsize$\pmb{\pm}$\textbf{0.14}} & 38.28{\scriptsize$\pm$0.72} & 38.41{\scriptsize$\pm$0.35} & 37.85{\scriptsize$\pm$0.24} & - \\ \hline
\makecell{SS \\ ($\text{mAP}^m$)} & 73.85{\scriptsize$\pm0.15$} & \textbf{74.68}{\scriptsize$\pmb{\pm}\textbf{0.37}$} & - & - & 74.49{\scriptsize$\pm0.43$} \\ \hline
\makecell{IS \\ (mIoU)} & 22.96{\scriptsize$\pm0.09$} & - & \textbf{23.72}{\scriptsize$\pmb{\pm}\textbf{0.16}$} & - & - \\ \hline
\makecell{D \\ (RMSE)} & 5.52{\scriptsize$\pm0.02$} & \textbf{5.26}{\scriptsize$\pmb{\pm}\textbf{0.01}$} & - & - & 5.38{\scriptsize$\pm0.02$} \\ \hline
\makecell{ID \\ ($l_1$ loss)} & 8.29{\scriptsize$\pm0.07$} & - & - & \textbf{8.25}{\scriptsize$\pmb{\pm}\textbf{0.27}$} & - \\ \hline
\end{tabular}}
\vspace*{1mm}
\caption{UniNet results on different task combinations: all five tasks together, single task (three of the five tasks, since the other two tasks IS and ID cannot be trained without object detection). Therefore, OD+IS and OD+ID results are reported as proxy single tasks. We also include the commonly used combination of SS+D.}
\label{table:all_tasks}
\end{table}

\subsection{Setup}
\textbf{Dataset}. The Cityscapes dataset \cite{Cordts2016Cityscapes} consists of images from driving scenes captured in European cities. The dataset consists of 2975, 500, and 1525 images in the training, validation and test sets, respectively. The images are of resolution 1024$\times$2048. The best fitting bounding box of each ground truth instance polygon is used for bounding box regression.

The NYUv2 \cite{Silberman2012IndoorSA} is an indoor dataset with image resolution of 480$\times$640 consisting of 795 and 654 training and validation images, respectively.

\textbf{Training details}. The encoder features E2 to E5 are obtained using DLA34 (Deep Layer Aggregation) \cite{Yu2018DeepLA} and the features E6 and E7 are obtained using VoVNet19 \cite{lee2019energy}. For all models, DLA34 is initialized with COCO \cite{Lin2014MicrosoftCC} pretrained weights unless otherwise mentioned. We train all combinations of multi-task models and single task models for a total of 140 epochs with a learning rate of 0.0001 and stepwise schedule where the learning rate is dropped by a factor of 10 at steps 98 and 126. The input images are resized to resolution $512\times1024$.

\textbf{Task evaluation}. Object detection and instance segmentation are evaluated at a resolution of $1024\times2048$. Semantic segmentation, depth estimation and instance depth prediction are evaluated at a resolution of $512\times1024$. For depth evaluation, we ignore all pixel locations where the ground truth depth is greater than 80 or less than 1e-3. All results are reported on the validation set for both datasets.

\textbf{Inference efficiency}. To determine inference efficiency, we measure the MAC, number of parameters, inference speed (FPS) and energy consumed by models. FPS and energy are reported on a single NVIDIA RTX-2080 Ti GPU. Following \cite{pi2017gpu} to measure inference energy, we run 500 forward passes with mini-batch size 1 of randomly generated "images" and report the average energy per image. FPS is also measured in a similar fashion using images from the validation set.

\textbf{Attack evaluation}. We use metric ratio to evaluate adversarial attacks. For any given task, metric ratio is the fraction of performance retained after adversarial attack with respect to the performance before attack.

\subsection{Multi-Task Performance}

Table \ref{table:sota} compares UniNet with state-of-the-art (SOTA) methods in both Cityscapes and NYUv2 datasets on SS+D task. The SOTA methods, the single task networks and the MTL-baseline are trained with the DLA34 backbone using the code provided by \cite{v2020revisiting}. Similar training hyperparameters are used across all methods. In both SS and D, MTI-Net achieves the best overall multi-task performance followed by UniNet. UniNet requires the least amount of computation (lowest MAC) and is also the fastest network (highest FPS). The energy consumption of UniNet is low and is only rivaled by PADNet and MTAN. However, both PADNet and MTAN lag behind UniNet in performance. Overall, UniNet can serve as an attractive baseline to study multi-task relationships due to its simplicity and optimum trade-off between performance and inference efficiency.

Table \ref{table:all_tasks} shows the results obtained using UniNet. As IS and ID cannot be trained alone, we consider the two task combinations OD+IS and OD+ID as proxy single tasks. Results for further task combinations are provided in Table S1. 

\subsection{Semantic and Geometric PGD Attacks}
\label{sec:pgdattacks}

\begin{figure*}
    \centering
        \begin{subfigure}{.39\textwidth}
          \centering
          \includegraphics[width=1\linewidth]{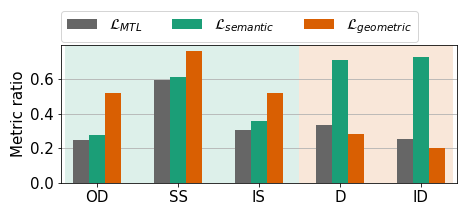}
          \caption{Multi-task, semantic and geometric loss}
          \label{fig:pgdgeoseg}
        \end{subfigure}%
        \begin{subfigure}{.61\textwidth}
          \centering
          \includegraphics[width=1\linewidth]{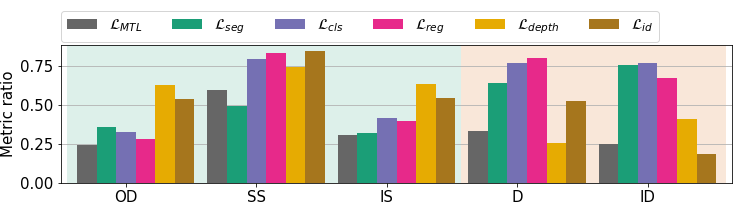}
          \caption{Multi-task and single task loss}
          \label{fig:pgdall}
        \end{subfigure}
   \caption{PGD attacks using different losses with $\epsilon$=1 as $l_\infty$ perturbation bound. Each bar represents a PGD attack conducted on UniNet with a specific loss (indicated with the bar color). The green shaded and red shaded region contains the metric ratios of semantic tasks and geometric tasks, respectively. Best viewed in color.}
\label{fig:pgd}
\end{figure*}

\begin{figure*}
    \centering
        \begin{subfigure}{1\textwidth}
          \centering
          \includegraphics[width=1\linewidth]{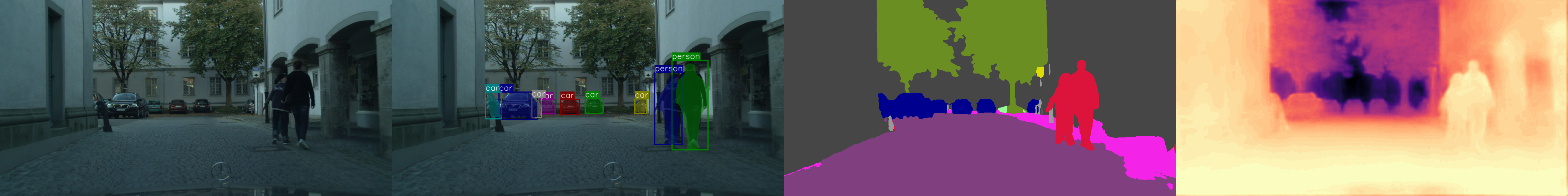}
          \caption{Clean image predictions}
          \label{fig:clean}
        \end{subfigure}%
        
        \begin{subfigure}{1\textwidth}
          \centering
          \includegraphics[width=1\linewidth]{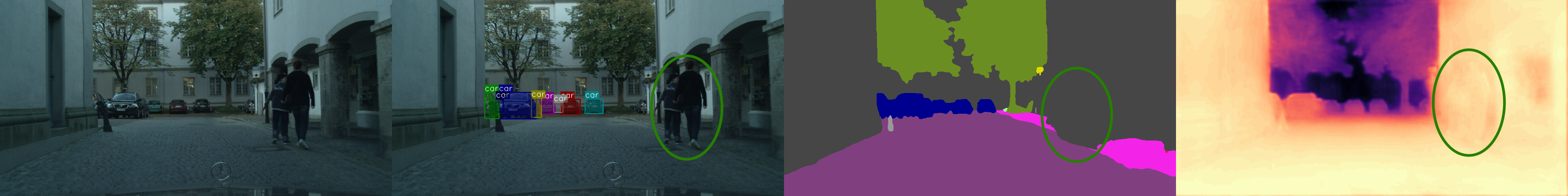}
          \caption{Hide ``persons" from segmentation map}
          \label{fig:segremove}
        \end{subfigure}%
        
        \begin{subfigure}{1\textwidth}
          \centering
          \includegraphics[width=1\linewidth]{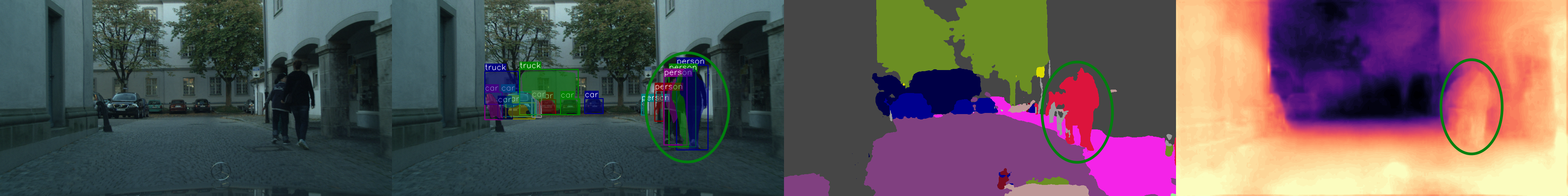}
          \caption{Hide ``persons" from depth map}
          \label{fig:depthremove}
        \end{subfigure}
   \caption{Semantic category hiding with $\epsilon$=2 as $l_\infty$ perturbation bound. (a) shows predictions on clean image. Hiding ``persons" from segmentation map in (b) also hides ``persons" from all task predictions. However, in (c) similar effect is not observed when hiding ``persons" from depth map. Starting from the left, the columns show the input image, the instance predictions, the predicted segmentation map and the predicted depth map. Best viewed in color.}
\label{fig:semanticremove}
\end{figure*}

PGD can be used to attack UniNet using different losses as discussed in Section \ref{sec:pgd}. We experiment with $l_\infty$ bound, varied $\epsilon$ values with step size $\alpha$=1. The number of attack iterations is determined by $\min (\epsilon+4,\lceil 1.25 \epsilon\rceil)$. In this section, we evaluate the effect of attacking with multi-task, semantic, and geometric losses. Given that the generated perturbation is expected to increase the attack loss, intuitively, the affected image features are important for the task which concerns the attack loss. Detailed results of adversarial attack on different task combinations and single tasks are provided in Table S2.


Figure \ref{fig:pgdgeoseg} highlights the metric ratios of all tasks under multi-task loss (MTL loss), semantic loss and geometric loss attacks. In the green region, semantic tasks OD, SS and IS are least affected by geometric loss. As indicated in the figure, semantic tasks retain most of their performance in relation to clean performance under geometric attack. Likewise, geometric tasks are least affected by semantic attacks (red region). This finding indicates that semantic and geometric tasks look at different image features.

\subsection{PGD Attacks with Individual Losses}

In this section, we analyze the effects of attacking with individual task losses as illustrated in Figure \ref{fig:pgdall}. MTL loss only affects object detection and instance segmentation more than their corresponding individual losses. This supports the notion that the MTL loss cannot effectively fool all tasks inline with Mao \etal \cite{Mao2020MultitaskLS}. Additional observations include (1) all tasks lose the most performance when attacked with the corresponding individual loss in comparison with other individual losses. (2) Semantic segmentation is the least affected task regardless of which loss is used. This indicates that semantic segmentation is the most robust task. We leave a more rigorous study of this observation as future work.


\subsection{Semantic and Geometric Task Relationships}

In Section \ref{sec:pgdattacks}, we observed with the aid of PGD attacks that the semantic and geometric tasks rely on different image features. In this section, we aim to study the relationships between semantic and geometric tasks. These tasks interact in the shared representation and this interaction is learned based on inter-task relationships during training. Semantic category hiding is a targeted attack in which the attack objective is formulated with the goal of hiding certain categories of objects from predictions. Given that the hiding can either be performed on the predicted segmentation map or the depth map, the effect of one prediction based attack on the other provides an opportunity to analyze semantic and geometric task relationships. 

Figure \ref{fig:semanticremove} shows the visualizations of predictions (a) before attack, (b) and (c) after attack to remove ``persons" from segmentation map (segmentation attack) and depth map (depth attack), respectively. In both attacks, the person region is filled using the surrounding building, road, and sidewalk pixels. The segmentation attack in Figure \ref{fig:segremove} is able to effectively hide ``persons" from all predictions, including depth estimation, which is a geometric task. However, the depth attack in Figure \ref{fig:depthremove} is able to blend ``persons" depth into the scene but doesn't completely affect semantic task predictions. This result shows that the semantic and geometric task do not affect each other equally.


\begin{table}
\centering
\small
\begin{tabular}{|l|c|cc|cc|}
\hline
\multirow{ 2}{*}{Tasks} & \multirow{ 2}{*}{Clean} & \multicolumn{ 2}{c|}{Hide from} & \multicolumn{ 2}{c|}{Hide from (rel.)} \\ \cline{3-6}
 &  & \makecell{SS} & \makecell{D} & \makecell{SS} & \makecell{D} \\ \hline
\makecell{OD ($\text{mAP}^b$)} & 35.23 & 8.95 & 13.85 & 0.25 & 0.39 \\ 
\makecell{IS ($\text{mAP}^m$)} & 16.46 & 2.83 & 5.38 & 0.17 & 0.33 \\ 
\makecell{SS (mIoU)} & 77.75 & 10.39 & 49.31 & \textcolor{blue}{0.13} & \textcolor{red}{0.63} \\ \hline 
\makecell{D (RMSE)} & 5.02 & 9.15 & 35.20 & \textcolor{red}{0.55} & \textcolor{blue}{0.14} \\ 
\makecell{ID (abs rel.)} & 0.16 & 0.29 & 1.05 & 0.54 & 0.15 \\ \hline
\end{tabular}
\vspace*{1mm}
\caption{Results on semantic hiding of ``person" class. The first three columns show the actual metric values while the last two columns show metric ratios which is the performance retained by each task after attack with respect to clean performance.}
\label{table:semhide}
\end{table}

\subsubsection{Role of Representation Levels}

To understand the role of representation levels in inter-task relationships, we analyze (i) hiding ``persons" from segmentation map (Figure \ref{fig:segremove}) and (ii) hiding ``persons" from depth map (Figure \ref{fig:depthremove}). We combine both analyses and present the final inference. 

(i) Figure \ref{fig:segremove} shows that the segmentation attack is able to fool all tasks resulting in ``persons" being hidden from all predictions. Given this observation, we infer that the segmentation attack likely disrupts all levels of representation. 

(ii) Figure \ref{fig:depthremove} shows that the depth attack is able to fool the geometric task. Both semantic segmentation and object detection are able to predict ``persons" in the right region but the shape of the ``person" masks and bounding boxes are affected. We infer that the depth attack has disrupted the low-level representation affecting both depth estimation and ``person" shape in semantic tasks. From this we infer that both semantic and geometric tasks strongly interact in low-level representation. However, at higher level representations, this interaction becomes weaker giving semantic tasks the room to still identify the class of ``persons".

In summary, (i) indicates that semantic and geometric tasks interact in all levels, and (ii) shows that the interaction is stronger in low-level representations. Thus, we conclude the semantic and geometric tasks have stronger interactions in low-level representations which becomes weaker in high-level representations. Furthermore, we observe that there is an asymmetric relationship between semantic and geometric tasks, that is the semantic tasks have stronger influence on geometric tasks in comparison to the influence of geometric tasks on semantic tasks.


The segmentation IoU of ``person" class and depth RMSE of ``person" regions in segmentation ground truth, before and after attack are shown in Table \ref{table:semhide}. Each of the attacks has the most effect on the corresponding task as is indicated by the numbers highlighted in blue. While depth estimation retains only 55\% of its performance under segmentation attack, semantic segmentation retains 63\% under depth attack. 





\subsection{Intra-task Relationship Induces Shape Bias}

We use the modified DAG attack discussed in Section \ref{sec:dagintro} to swap ``person" and ``car" classes in the bounding box predictions of object detection. We analyze the intra-task relationship in object detection between the classification (\textbf{cls}) and regression (\textbf{reg}) objectives. Notably, the two objectives are handled by two different branches in the UniNet.

\begin{figure}
    \centering
    \includegraphics[width=0.85\linewidth]{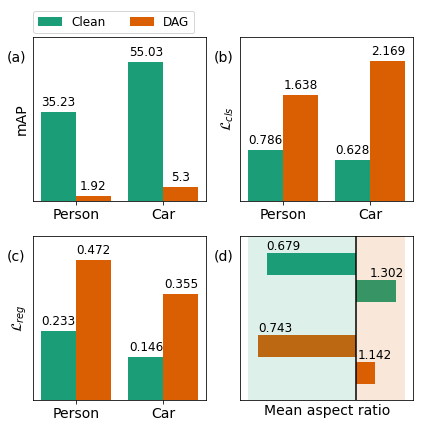}
   \caption{Effect of DAG attack on object detection. Each plot compares before and after DAG attack results specific to the ``person" and ``car" class. Plots (a), (b) and (c) show the mAP, classification loss and regression loss, respectively. Plot (d) shows the mean aspect ratio of ``person" class (green shaded region) and ``car" class (red shaded region).}
\label{fig:shapemetrics}
\end{figure}

In the case where there is no relationship between \textbf{cls} and \textbf{reg}, the DAG attack can be expected to only affect the class of bounding boxes. The results before and after DAG attack is presented in Figure \ref{fig:shapemetrics}. Plot (a) shows that DAG attack is effective as the ``person" and ``car" mAP has dropped considerably relative to clean performance. In plot (b), we see that the classification loss has increased as expected. However, the regression loss in plot (c) also increases suggesting that DAG has exploited the learned interaction between \textbf{cls} and \textbf{reg} which is a result of their relationship.

\begin{figure}
    \centering
        \begin{subfigure}{.4\textwidth}
          \centering
          \includegraphics[width=0.95\linewidth]{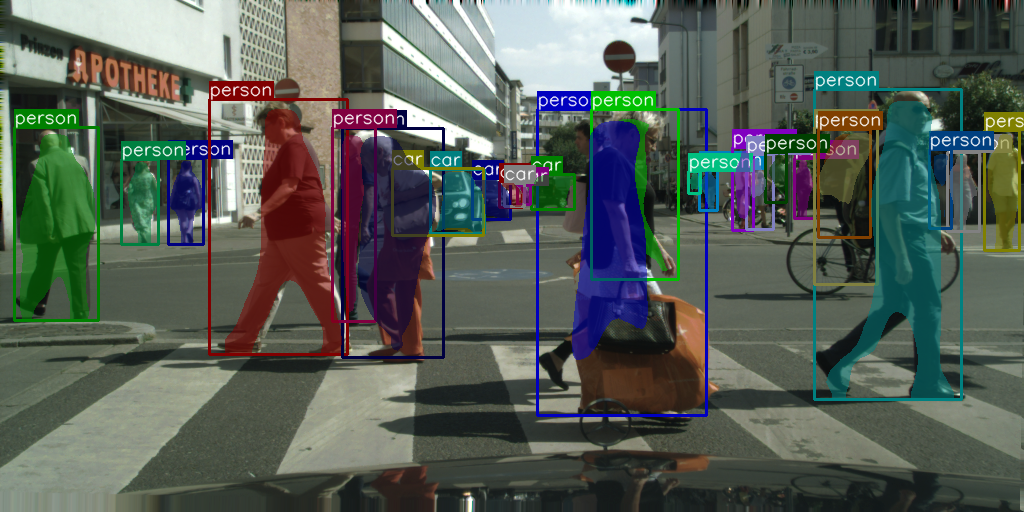}
          \caption{Predictions on clean image.}
          \label{fig:biasorig}
        \end{subfigure}
        
        \begin{subfigure}{.4\textwidth}
          \centering
          \includegraphics[width=0.95\linewidth]{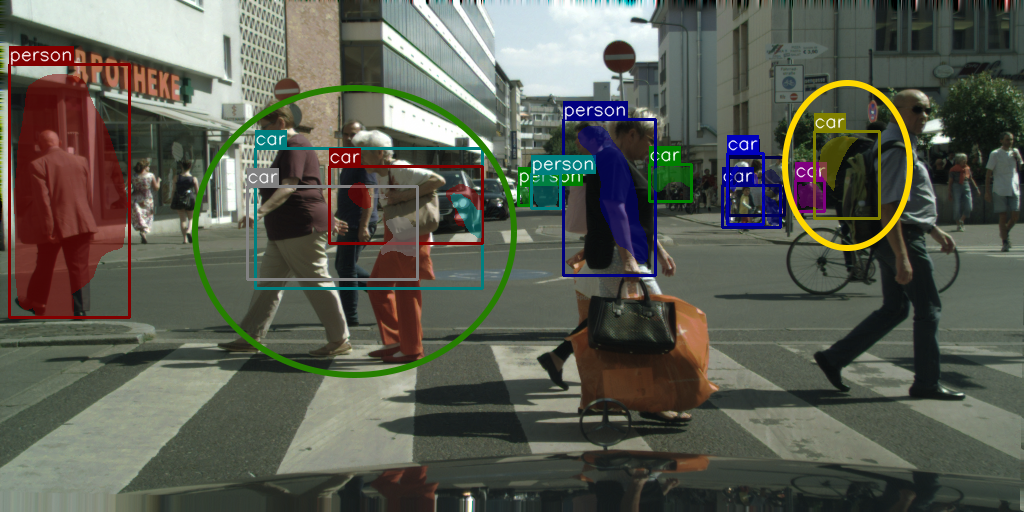}
          \caption{Predictions on DAG generated adversarial image.}
          \label{fig:biasadv}
        \end{subfigure}
   \caption{Shape bias induced by intra-task relationship in object detection. OD and IS predictions visualized on (a) clean image and (b) DAG generated adversarial image. In (b), ``car" box predictions on ``persons" have horizontal aspect ratio despite that the attack objective only includes box classification. The visualization only includes ``car" and ``person" predictions. Best viewed in color.}
\label{fig:shapebias}
\end{figure}

In the Cityscapes dataset, ``car" boxes are frequently horizontal or approximately square. Figure \ref{fig:shapebias} shows the visualisation of instance predictions (a) on clean image and (b) DAG attacked image. We see that the ``person" bounding boxes have switched from vertically oriented boxes to either horizontally oriented (green oval region) or approximately square boxes (yellow oval region). This observation suggests that DAG has exploited a shape bias in the learned representation. This shape bias has likely been induced by the relationship between \textbf{cls} and \textbf{reg} leveraging a similar bias in the dataset during training. As ``cars" can generally be expected to have such shapes in the real world, this shape bias is desirable.

To further evaluate the observed shape bias, we consider the mean aspect ratio of predicted ``person" and ``car" bounding boxes across the validation set as a proxy indicator of shape. Figure \ref{fig:shapemetrics} plot (d) shows the mean aspect ratios of ``person" and ``car" predictions on clean and DAG attacked images. Given that the attack only swaps the two classes, one would expect the mean aspect ratio after attack to switch regions. However, we see that ``person" and ``car" mean aspect ratios remain below 1 and above 1, respectively. This shows that the aspect ratios have also swapped along with the class, reinforcing the finding that a shape bias exists in the learned representation induced by the intra-task interaction between \textbf{cls} and \textbf{reg}.

\subsection{Ablation study}
To check whether the findings are architecture independent, we choose MTI-Net because (i) it has the best performance in segmentation and depth, and (ii) its architecture is quite different from UniNet. We add the instance head to MTI-Net and refer to this new architecture as MTI-Net++. This network is then attacked using all three adversarial attacks revealing findings in line with that obtained using UniNet. These results are provided in the supplementary material and suggest that the conclusions made regarding multi-task relationships are architecture independent.




\section{Conclusions}
We presented an efficiently designed unified scene understanding network, UniNet. We demonstrated that it has competitive performance with existing works and provides a comprehensive understanding of a scene with the aid of five vision tasks. We introduced adversarial attacks as an exploratory lens to understand and obtain insights about multi-task relationships. With the aid of semantic category hiding, we showed that semantic and geometric tasks have an asymmetric relationship meaning that semantic tasks have a stronger effect on geometric tasks. We also showed that their strong interaction in low-level representations becomes weaker as we move towards high-level representations. With targeted DAG attack, we studied the effects of swapping two instance classes in object detection and showed that intra-task relationship between classification and regression induces a desirable bias towards object shapes in the learned representation. Adversarial attacks thus provide an interesting front to study multi-task relationships.

{\small
\bibliographystyle{ieee_fullname}
\bibliography{egbib}
}

 \newpage

\section{Multi-task combinations}

We trained UniNet with all valid task combinations and provided the results in Table \ref{table:SM_combs}. The divisions in the table are based on the number of tasks in the combination. For instance, the third division provides results of three task combinations. For each task, we highlight the best performance across divisions. Except for depth estimation, all tasks achieve their best performance when learned jointly with the other tasks.

\begin{table*}
    \centering
    \small
    \begin{tabular}{|c|ccc|ccc|}
    \hline
    \multirow{ 2}{*}{Tasks} & OD & SS & IS & \multicolumn{2}{c}{D}  & ID \\ \cline{2-7}
     &  $\text{mAP}^b$ & mIoU & $\text{mAP}^m$ & RMSE & Abs err. & $l_1$ loss \\ \hline 
    
    OD+SS+IS+D+ID & 38.93{\scriptsize$\pm$0.14} & 73.85{\scriptsize$\pm0.15$} & 22.96{\scriptsize$\pm0.09$} & 5.52{\scriptsize$\pm0.02$} & 2.91{\scriptsize$\pm0.03$} & 8.29{\scriptsize$\pm0.07$}  \\ \hline 
    
    OD+SS+D+ID & 38.45{\scriptsize$\pm$0.48} & 73.54{\scriptsize$\pm0.49$} & - & 5.51{\scriptsize$\pm0.01$} & 2.89{\scriptsize$\pm0.02$} & 8.38{\scriptsize$\pm0.06$}  \\ 
    
    OD+SS+IS+D & 39.11{\scriptsize$\pmb{\pm}$0.80} & 74.72{\scriptsize$\pm0.46$} & \textbf{23.90}{\scriptsize$\pmb{\pm}\textbf{0.80}$} & 5.45{\scriptsize$\pm0.04$} & 2.87{\scriptsize$\pm0.03$} & -  \\  
    
    OD+SS+IS+ID & 38.35{\scriptsize$\pm$0.32} & 74.15{\scriptsize$\pm0.80$} & 22.75{\scriptsize$\pm0.27$} & - & - & 8.428{\scriptsize$\pm0.102$}  \\ 
    
    OD+IS+D+ID & 38.15{\scriptsize$\pm$0.38} & - & 22.84{\scriptsize$\pm0.34$} & 5.443{\scriptsize$\pm0.033$} & 2.83{\scriptsize$\pm0.03$} & 8.10{\scriptsize$\pm0.10$}  \\ \hline
    
    OD+SS+D & \textbf{39.22{\scriptsize$\pm$0.06}} & 74.98{\scriptsize$\pm0.06$} & - & 5.45{\scriptsize$\pm0.02$} & 2.89{\scriptsize$\pm0.02$} & -  \\ 
    
    OD+SS+ID & 38.52{\scriptsize$\pm$0.61} & 74.42{\scriptsize$\pm0.22$} & - & - & - & 8.41{\scriptsize$\pm0.11$}  \\ 
    
    OD+SS+IS & 39.01{\scriptsize$\pm$0.25} & \textbf{75.07}{\scriptsize$\pmb{\pm}\textbf{0.49}$} & 23.72{\scriptsize$\pm0.07$} & - & - & -  \\ 
    
    OD+D+ID & 38.12{\scriptsize$\pm$0.64} & - & - & 5.46{\scriptsize$\pm0.04$} & 2.86{\scriptsize$\pm0.01$} & \textbf{8.03}{\scriptsize$\pmb{\pm}\textbf{0.11}$}  \\ 
    
    OD+IS+D & 39.08{\scriptsize$\pm$0.58} & - & 23.72{\scriptsize$\pm0.28$} & 5.34{\scriptsize$\pm0.03$} & 2.75{\scriptsize$\pm0.01$} & -  \\ 
    
    OD+IS+ID & 38.29{\scriptsize$\pm$0.60} & - & 22.64{\scriptsize$\pm0.15$} & - & - & 8.16{\scriptsize$\pm0.10$}  \\ \hline
    
    OD+SS & 38.22{\scriptsize$\pm$0.53} & 75.06{\scriptsize$\pm0.10$} & - & - & - & -  \\ 
    
    OD+D & 38.50{\scriptsize$\pm$0.62} & - & - & 5.34{\scriptsize$\pm0.03$} & 2.76{\scriptsize$\pm0.02$} & -  \\ 
    
    OD+ID & 37.85{\scriptsize$\pm$0.24} & - & - & - & - & 8.25{\scriptsize$\pm0.27$}  \\ 
    
    OD+IS & 38.41{\scriptsize$\pm$0.35} & - & 23.72{\scriptsize$\pm0.16$} & - & - & -  \\ 
    
    SS+D & - & 74.49{\scriptsize$\pm0.43$} & - & 5.38{\scriptsize$\pm0.02$} & 2.81{\scriptsize$\pm0.01$} & -  \\ \hline
    
    OD & 38.28{\scriptsize$\pm$0.72} & - & - & - & - & -  \\ 
    
    SS & - & 74.68{\scriptsize$\pm0.37$} & - & - & - & -  \\ 
    
    D & - & - & - & \textbf{5.26}{\scriptsize$\pmb{\pm}\textbf{0.01}$} & \textbf{2.70}{\scriptsize$\pmb{\pm}\textbf{0.02}$} & -  \\ \hline
    
    \end{tabular}
    \vspace*{0.5mm}
    \caption{Results of all possible task combinations. For all tasks except depth estimation, the best performance is obtained in one of the multi-task combinations.}
    \label{table:SM_combs}
\end{table*}

\section{PGD Attacks}

Table \ref{table:SM_pgd} provides the results of PGD attack on the five task UniNet model and single task models. The single task model attacks are performed using the corresponding task loss ($\mathcal{L}_{task}$).

\begin{table}
	\centering
	\resizebox{\columnwidth}{!}{%
	\begin{tabular}{|c|c|ccc|cc|}
	\hline
	\multirow{ 2}{*}{\makecell[l]{Loss used \\ for attack}} & \multirow{ 2}{*}{$\epsilon$} & OD & SS & IS & D & ID \\ \cline{3-7}
	 & & $\text{mAP}^b$ & mIoU & $\text{mAP}^m$ & RMSE & $l_1$ loss \\ \hline 
      
	 \multirow{ 5}{*}{$\mathcal{L}_{MTL}$} & 0.25 & 23.65 & 59.25 & 15.23 & 10.45 & 17.29 \\ 
	 & 0.5 & 18.91 & 55.76 & 12.79 & 11.87 & 20.34 \\ 
	 & 1 & 9.71 & 43.97 & 7.12 & 16.30 & 32.47 \\ 
	 & 2 & 5.94 & 35.19 & 4.51 & 19.62 & 45.60 \\ 
	 & 4 & 2.48 & 23.12 & 2.12 & 23.31 & 67.73 \\ \hline
     
	 \multirow{ 5}{*}{$\mathcal{L}_{semantic}$} & 0.5 & 21.59 & 58.95 & 14.83 & 6.20 & 9.58 \\ 
	 & 0.5 & 17.28 & 55.22 & 12.06 & 6.60 & 10.41 \\ 
	 & 1 & 10.77 & 45.21 & 8.31 & 7.73 & 11.33 \\ 
	 & 2 & 7.90 & 37.63 & 6.26 & 9.10 & 13.40 \\ 
	 & 4 & 4.29 & 26.61 & 3.41 & 11.53 & 16.07 \\ \hline
     
	 \multirow{ 5}{*}{$\mathcal{L}_{geometric}$} & 0.25 & 33.57 & 68.75 & 19.99 & 11.78 & 18.70 \\ 
	 & 0.5 & 28.81 & 64.77 & 17.15 & 13.55 & 22.35 \\ 
	 & 1 & 20.35 & 56.01 & 11.95 & 19.52 & 40.32 \\ 
	 & 2 & 14.00 & 46.38 & 8.09 & 24.06 & 62.75 \\ 
	 & 4 & 6.25 & 34.25 & 3.21 & 30.03 & 105.44 \\ \hline
       
	\multirow{ 5}{*}{$\mathcal{L}_{seg}$} & 0.25 & 26.95 & 53.47 & 15.30 & 6.46 & 8.77 \\ 
	 & 0.5 & 22.69 & 49.36 & 12.96 & 6.914 & 9.31 \\ 
	 & 1 & 14.15 & 36.45 & 7.47 & 8.54 & 10.83 \\ 
	 & 2 & 10.61 & 26.83 & 5.74 & 10.56 & 13.11 \\ 
	 & 4 & 5.99 & 13.77 & 3.07 & 15.02 & 17.30 \\ \hline
     
	 \multirow{ 5}{*}{$\mathcal{L}_{cls}$} & 0.25 & 23.45 & 67.48 & 15.82 & 6.06 & 9.32 \\ 
	 & 0.5 & 19.20 & 63.95 & 13.76 & 6.41 & 10.01 \\ 
	 & 1 & 12.82 & 58.91 & 9.59 & 7.14 & 10.67 \\ 
	 & 2 & 9.45 & 53.92 & 7.38 & 8.11 & 12.65 \\ 
	 & 4 & 6.40 & 48.92 & 5.14 & 9.37 & 14.89 \\ \hline
     
	 \multirow{ 5}{*}{$\mathcal{L}_{reg}$} & 0.25 & 24.13 & 70.10 & 16.87 & 5.98 & 9.62 \\ 
	 & 0.5 & 19.41 & 67.09 & 14.18 & 6.29 & 10.43 \\ 
	 & 1 & 11.16 & 61.75 & 9.22 & 6.84 & 12.17 \\ 
	 & 2 & 6.61 & 55.50 & 5.84 & 7.75 & 14.71 \\ 
	 & 4 & 2.22 & 47.23 & 2.22 & 9.07 & 18.54 \\ \hline
     
	 \multirow{ 5}{*}{$\mathcal{L}_{is}$} & 0.25 & 28.40 & 69.66 & 16.48 & 5.95 & 9.57 \\ 
	 & 0.5 & 24.02 & 66.55 & 13.80 & 6.28 & 10.33 \\ 
	 & 1 & 18.45 & 62.76 & 10.60 & 6.92 & 11.58 \\ 
	 & 2 & 13.78 & 57.91 & 7.69 & 7.80 & 13.59 \\ 
	 & 4 & 9.19 & 51.08 & 4.66 & 9.21 & 16.12 \\ \hline
     
	 \multirow{ 5}{*}{$\mathcal{L}_{depth}$} & 0.25 & 35.08 & 68.46 & 20.70 & 12.65 & 11.15 \\ 
	 & 0.5 & 31.41 & 64.90 & 18.47 & 14.62 & 13.09 \\ 
	 & 1 & 24.71 & 54.91 & 14.64 & 21.32 & 20.16 \\ 
	 & 2 & 17.20 & 42.07 & 10.14 & 27.02 & 27.91 \\ 
	 & 4 & 8.17 & 26.03 & 4.95 & 40.33 & 35.95 \\ \hline
     
	 \multirow{ 5}{*}{$\mathcal{L}_{id}$} & 0.25 & 32.88 & 70.31 & 19.59 & 7.15 & 20.63 \\ 
	 & 0.5 & 28.23 & 66.75 & 16.79 & 7.99 & 24.81 \\ 
	 & 1 & 21.02 & 62.46 & 12.60 & 10.45 & 44.49 \\ 
	 & 2 & 14.29 & 53.95 & 8.56 & 13.54 & 67.08 \\ 
	 & 4 & 6.85 & 41.71 & 3.77 & 16.38 & 108.60 \\ \hline
	 
	 \multirow{ 5}{*}{$\mathcal{L}_{task}$} & 0.25 & 20.80 & 55.59 & - & 11.81 & - \\
     & 0.5 & 17.33 & 51.05 & - & 13.71 & - \\
     & 1 & 10.31 & 36.99 & - & 20.59 & - \\
     & 2 & 6.86 & 27.52 & - & 26.78 & - \\
     & 4 & 4.17 & 14.01 & - & 38.71 & - \\ \hline
     
	\end{tabular}}
	\vspace*{1mm}
	\caption{PGD attack results with varying $\epsilon$ values on the five task and single task UniNet models using different losses.}
	\label{table:SM_pgd}
\end{table}

\section{Semantic category hiding}

In this section, we provide visualizations of more predictions obtained using the semantic category hiding attack. In addition to using the ``person" class for attack, we also use the ``car" class. The relevance of these results with respect to the findings of this work are discussed followed by some cases where the findings do not hold.

\begin{figure*}
    \centering
        \begin{subfigure}{1\textwidth}
          \centering
          \includegraphics[width=1\linewidth]{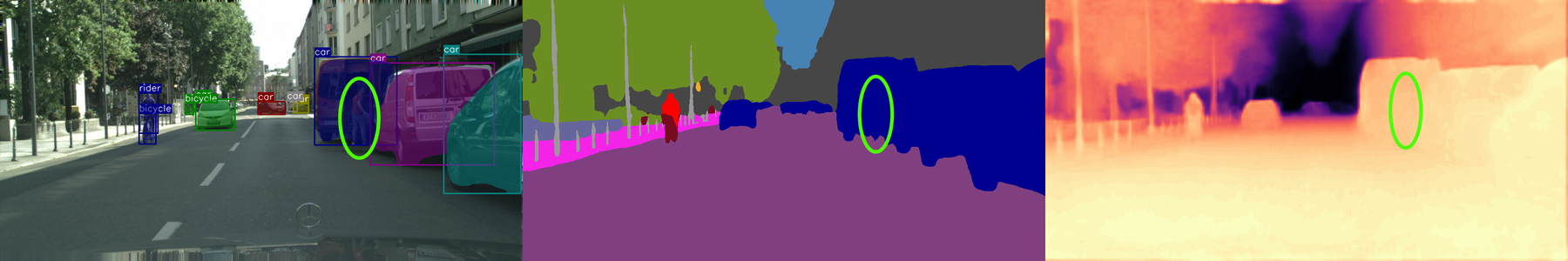}
        \end{subfigure}%
        
        \begin{subfigure}{1\textwidth}
          \centering
          \includegraphics[width=1\linewidth]{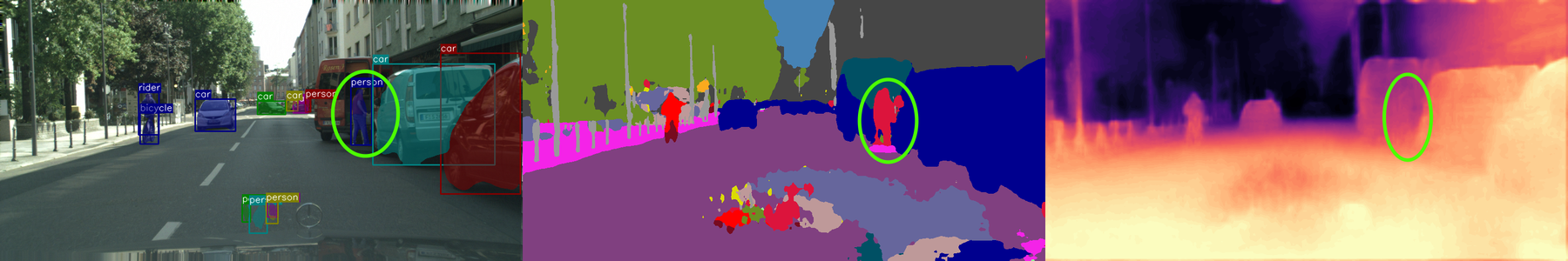}
          \caption{Predictions on first image.}
        \end{subfigure}
        
        \vspace*{4mm}
        \begin{subfigure}{1\textwidth}
          \centering
          \includegraphics[width=1\linewidth]{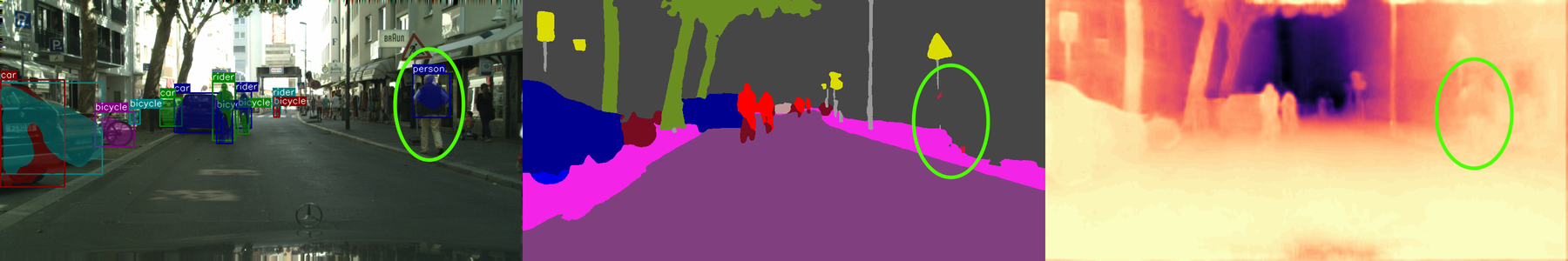}
        \end{subfigure}%
        
        \begin{subfigure}{1\textwidth}
          \centering
          \includegraphics[width=1\linewidth]{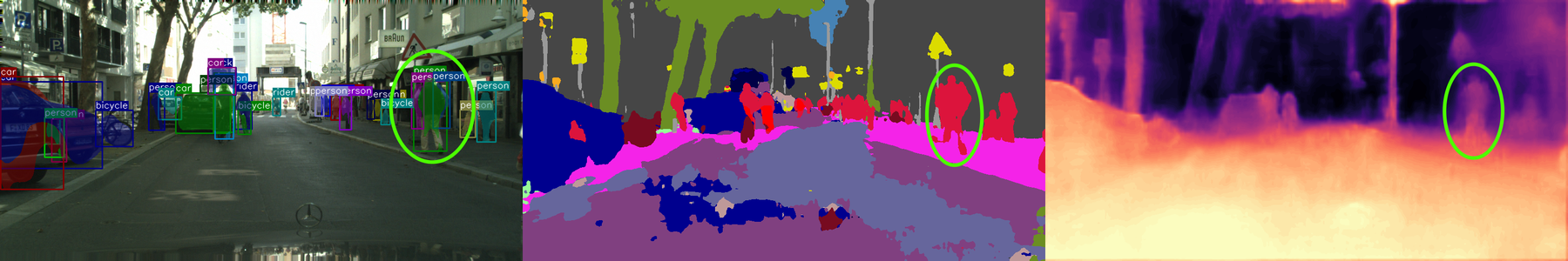}
          \caption{Predictions on second image.}
        \end{subfigure}
        
        \vspace*{4mm}
        \begin{subfigure}{1\textwidth}
          \centering
          \includegraphics[width=1\linewidth]{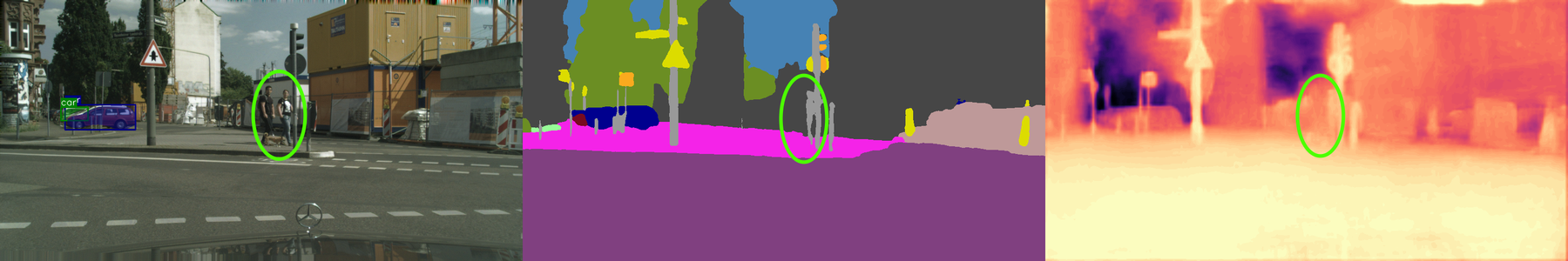}
        \end{subfigure}%
        
        \begin{subfigure}{1\textwidth}
          \centering
          \includegraphics[width=1\linewidth]{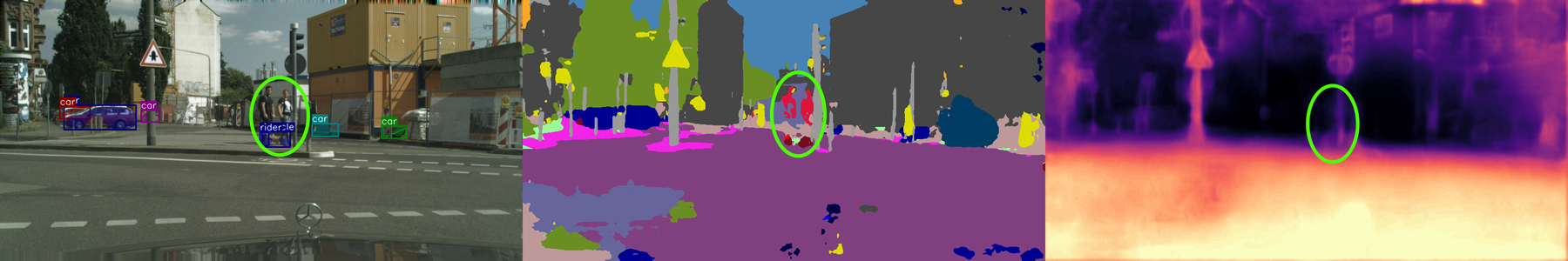}
          \caption{Predictions on third image.}
        \end{subfigure}
        
  \vspace*{1mm}
  \caption{Semantic category hiding of ``persons". Only the prediction visualizations are shown as changes in adversarial images are imperceptible for the used perturbation bound $\epsilon$=2. In each pairs of visualizations, the first and second row show hiding from segmentation and depth map results, respectively. Best viewed in color.}
\label{fig:SM_hideperson}
\end{figure*}

\begin{figure*}
    \centering
        \begin{subfigure}{1\textwidth}
          \centering
          \includegraphics[width=1\linewidth]{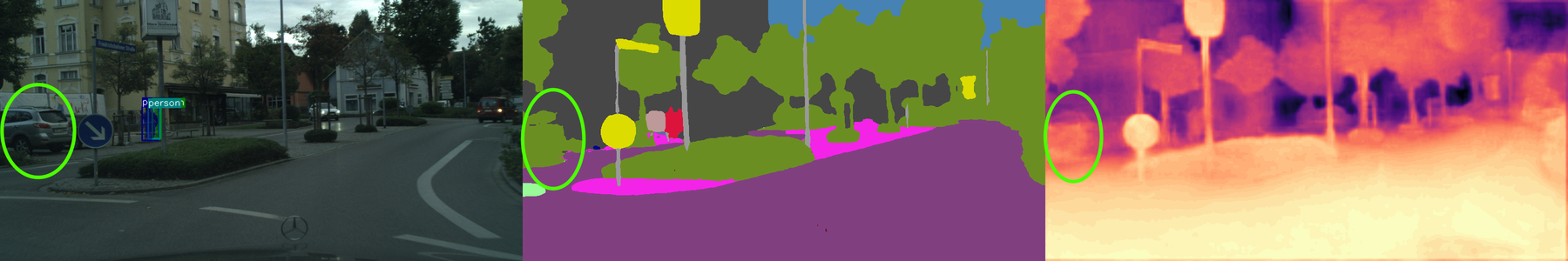}
        \end{subfigure}%
        
        \begin{subfigure}{1\textwidth}
          \centering
          \includegraphics[width=1\linewidth]{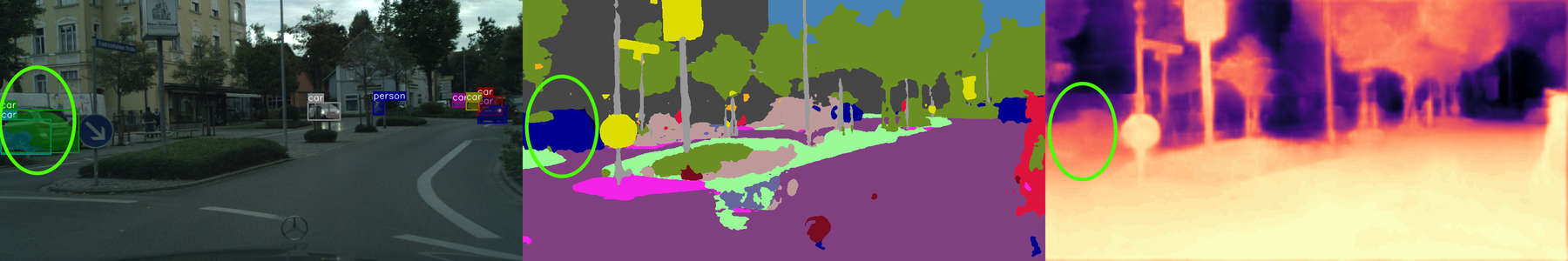}
          \caption{Predictions on first image.}
        \end{subfigure}
        
        \vspace*{4mm}
        \begin{subfigure}{1\textwidth}
          \centering
          \includegraphics[width=1\linewidth]{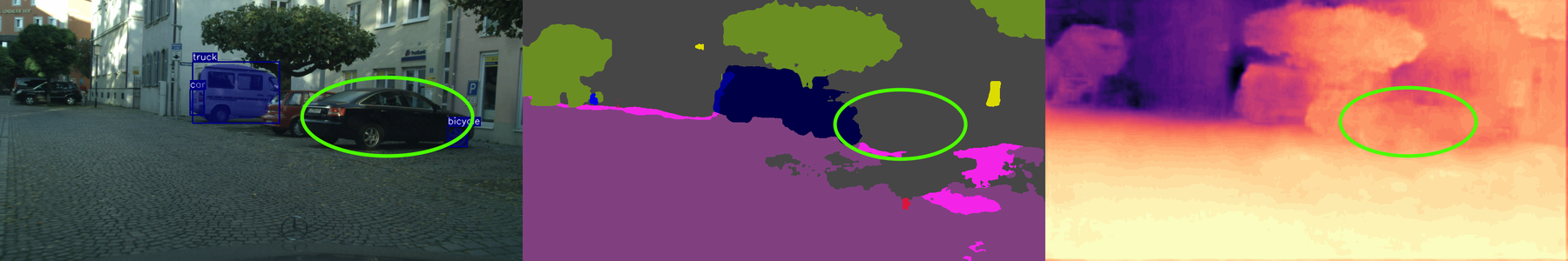}
        \end{subfigure}%
        
        \begin{subfigure}{1\textwidth}
          \centering
          \includegraphics[width=1\linewidth]{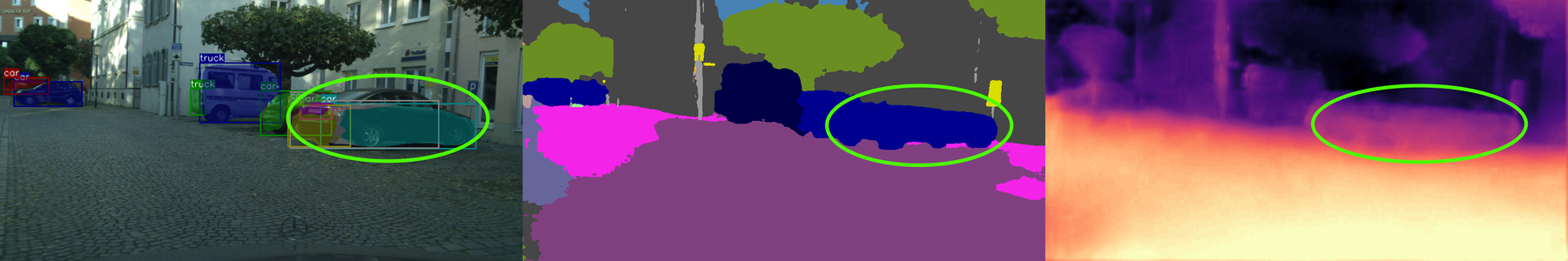}
          \caption{Predictions on second image.}
        \end{subfigure}
        
        \vspace*{4mm}
        \begin{subfigure}{1\textwidth}
          \centering
          \includegraphics[width=1\linewidth]{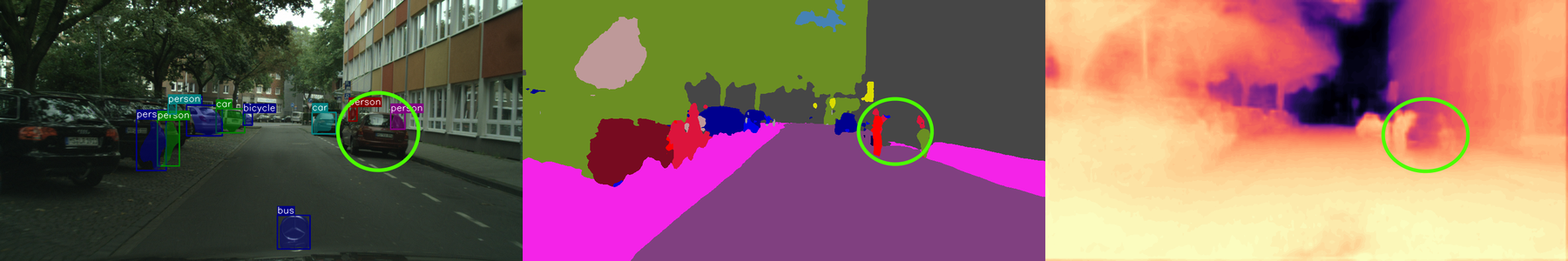}
        \end{subfigure}%
        
        \begin{subfigure}{1\textwidth}
          \centering
          \includegraphics[width=1\linewidth]{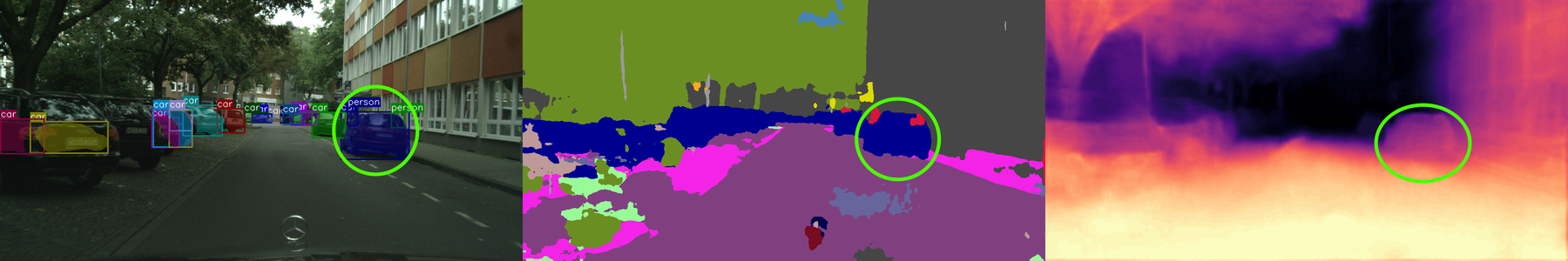}
          \caption{Predictions on third image.}
        \end{subfigure}
        
  \vspace*{1mm}
  \caption{Semantic category hiding of ``cars". Only the prediction visualizations are shown as changes in adversarial images are imperceptible at the used perturbation bound $\epsilon$=2. In each pairs of visualizations, the first and second row show hiding from segmentation and depth map results, respectively. Best viewed in color.}
\label{fig:SM_hidecar}
\end{figure*}

In Figure \ref{fig:SM_hideperson}, the predictions obtained by attacking the ``person" class is visualized. Predictions on three images are visualized in (a), (b) and (c). For all three images, the first row and the second row show the predictions when hiding ``persons" from segmentation map (segmentation attack) and depth map (depth attack), respectively. We present discussions based on the green oval regions. In all three images, segmentation attack results in ``persons" being removed from both the segmentation and depth predictions. However, in (b) we see that a few pixels still predict ``persons" correctly. Correspondingly, object detection also predicts ``person" but with an incorrect bounding box. In general, we note that the segmentation attack is able to affect all task predictions. In all three image predictions, ``persons" blend into the background when depth attack is used. The predictions in semantic tasks remain but their shapes are affected. In Figure \ref{fig:SM_hidecar}, we similarly provide results on the attacks regarding hiding ``cars". All the predictions follow the same trend like hiding ``persons". These results provide further support that the semantic and geometric tasks are asymmetric in the sense semantic tasks have stronger effect on geometric tasks in comparison to the effect of geometric tasks on semantic tasks. 

\begin{figure*}
    \centering
        \begin{subfigure}{1\textwidth}
          \centering
          \includegraphics[width=1\linewidth]{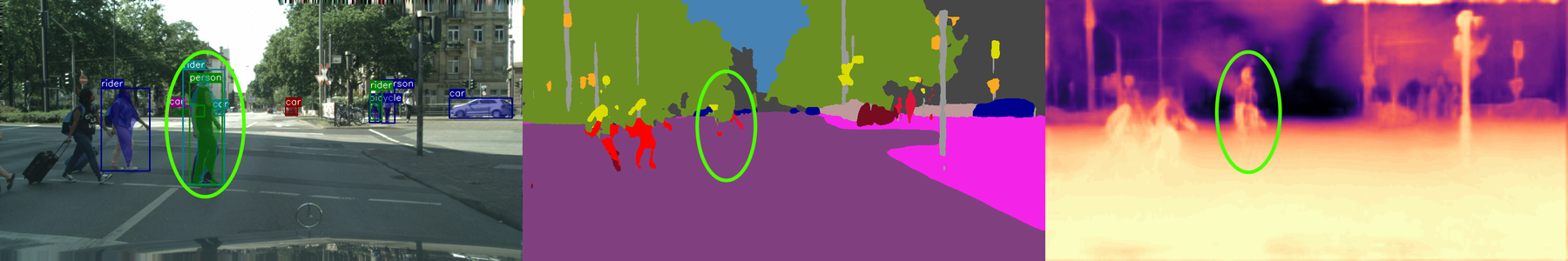}
        \end{subfigure}%
        
        \begin{subfigure}{1\textwidth}
          \centering
          \includegraphics[width=1\linewidth]{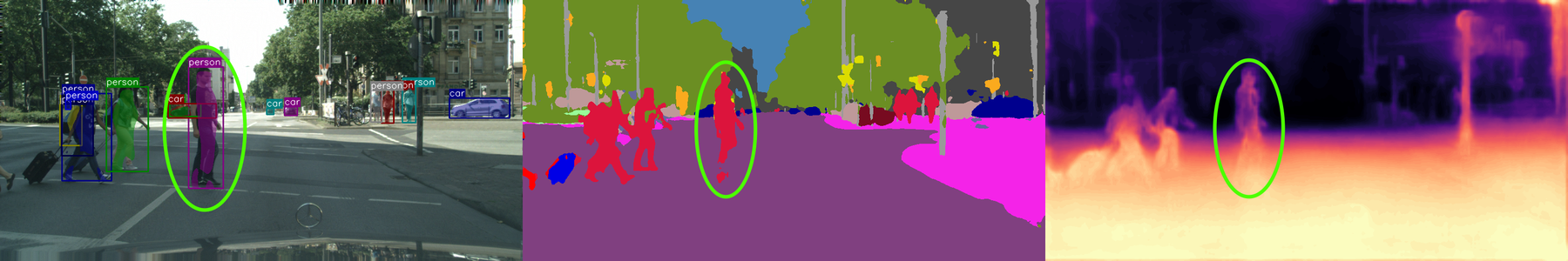}
          \caption{Predictions on first image.}
        \end{subfigure}
        
        \vspace*{4mm}
        \begin{subfigure}{1\textwidth}
          \centering
          \includegraphics[width=1\linewidth]{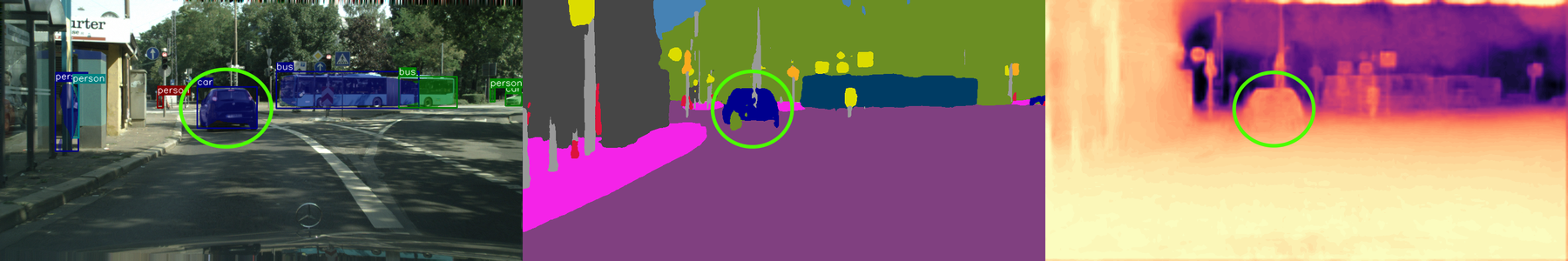}
        \end{subfigure}%
        
        \begin{subfigure}{1\textwidth}
          \centering
          \includegraphics[width=1\linewidth]{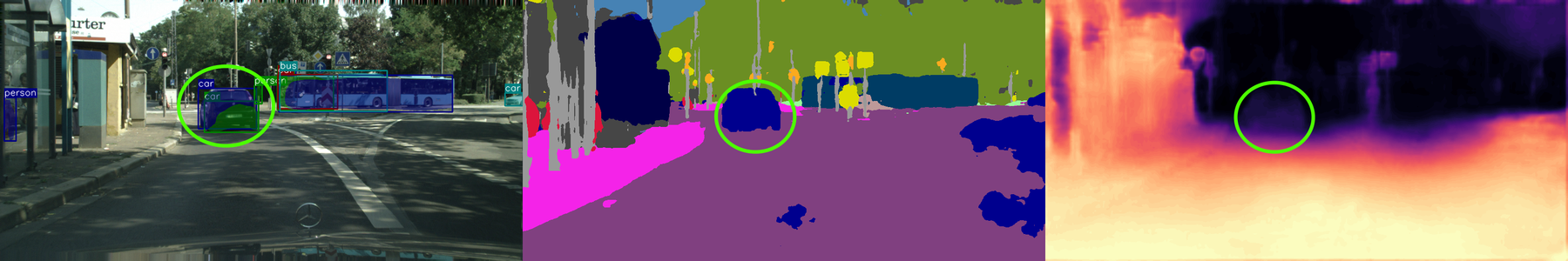}
          \caption{Predictions on second image.}
        \end{subfigure}
        
  \vspace*{1mm}
  \caption{Failure cases: the perturbation bound used is $\epsilon$=2. In each pairs of rows (divided by white space), the first and second row show hiding from segmentation and depth map results, respectively. Best viewed in color.}
\label{fig:SM_hidefailure}
\end{figure*}

In Figure \ref{fig:SM_hidefailure}, we show cases where the segmentation attack is not effective in hiding the intended class from the predictions. In the first row in (a) under segmentation attack, remnants of ``person" pixels still remain and object detection does predict ``person" correctly. Likewise, in (b) first row, the ``car" is not removed by the segmentation attack in both object detection and segmentation predictions. However, we note that segmentation attack is able to affect depth predictions while the depth attack only has minor effects on semantic predictions. 

\section{DAG attacks}

DAG can be used to attack both object detection and semantic segmentation in a targeted manner to swap ``person" and ``car" classes. We present more visualizations of predictions while attacking object detection with DAG. In addition, we also visualize predictions when semantic segmentation is attacked. The five task UniNet model is used for DAG attacks on both cases.

\begin{figure*}
    \centering
        \begin{subfigure}{1\textwidth}
          \centering
          \includegraphics[width=0.8\linewidth]{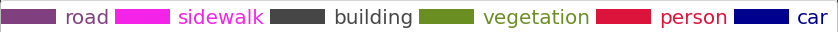}
        \end{subfigure}
        
        \vspace{1mm}
        \begin{subfigure}{0.45\textwidth}
          \centering
          \includegraphics[width=1\linewidth]{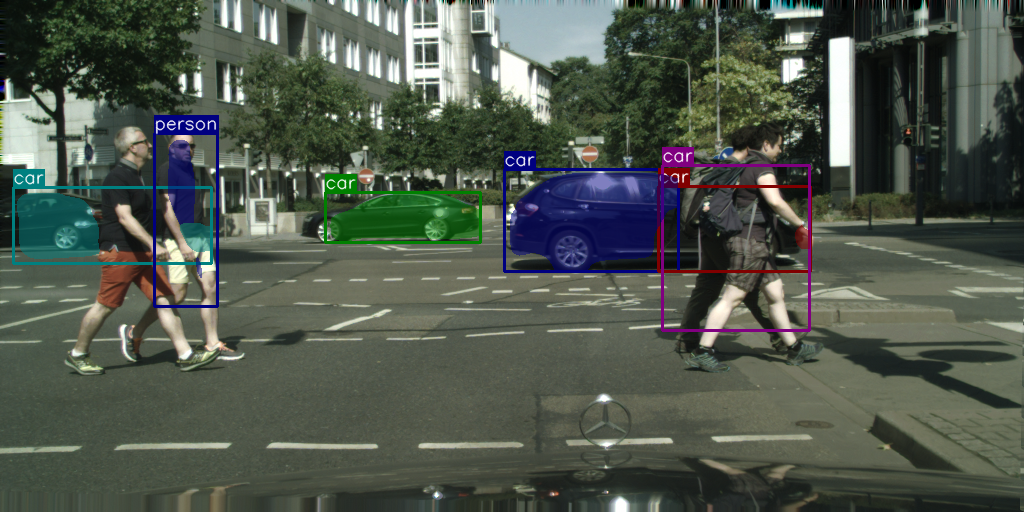}
        \end{subfigure}%
        \hspace*{0.5mm}
        \begin{subfigure}{0.45\textwidth}
          \centering
          \includegraphics[width=1\linewidth]{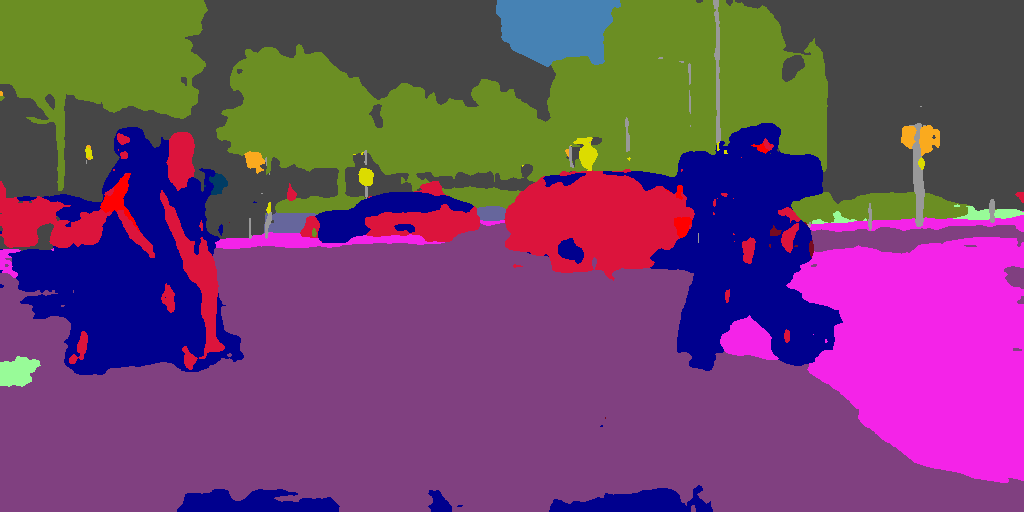}
        \end{subfigure}
        
        \vspace{2mm}
        \begin{subfigure}{0.45\textwidth}
          \centering
          \includegraphics[width=1\linewidth]{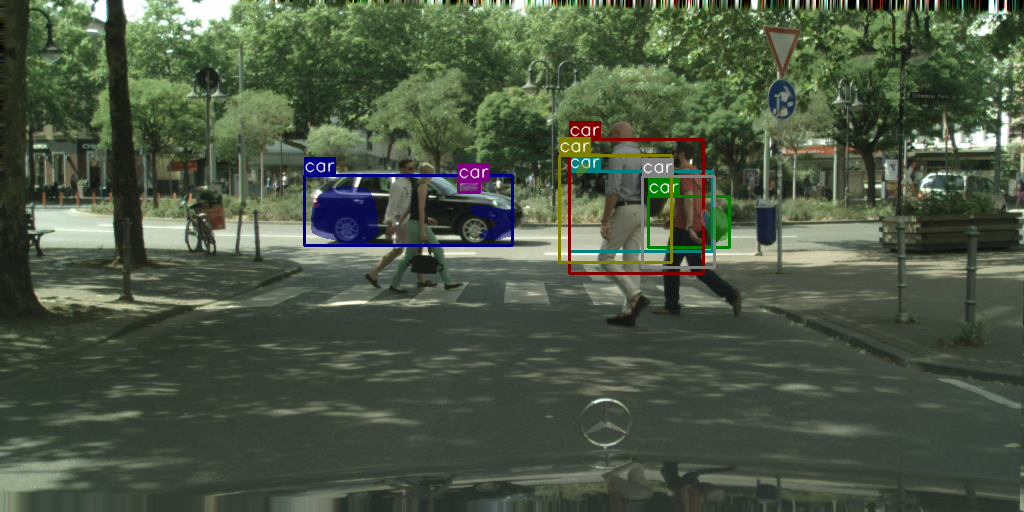}
        \end{subfigure}%
        \hspace*{0.5mm}
        \begin{subfigure}{0.45\textwidth}
          \centering
          \includegraphics[width=1\linewidth]{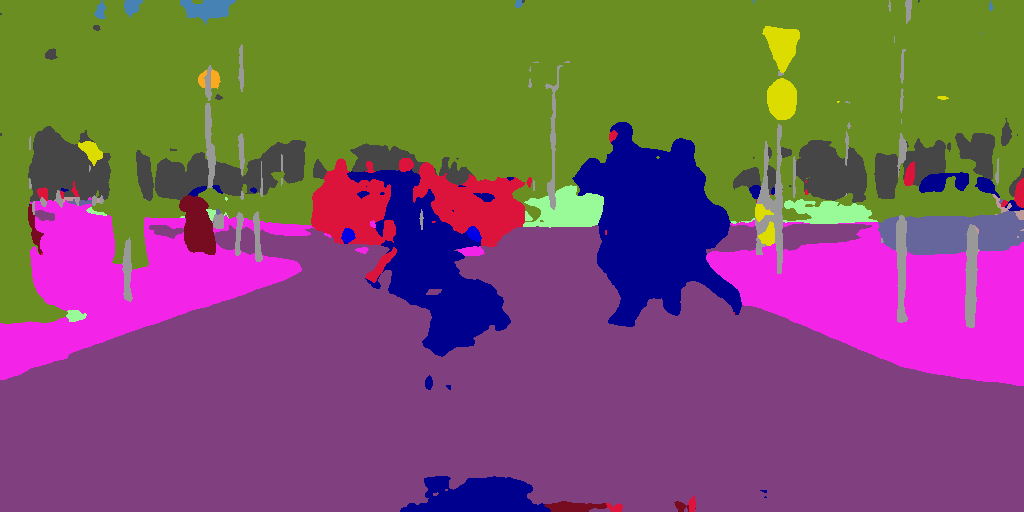}
        \end{subfigure}%
        
        \vspace{2mm}
        \begin{subfigure}{0.45\textwidth}
          \centering
          \includegraphics[width=1\linewidth]{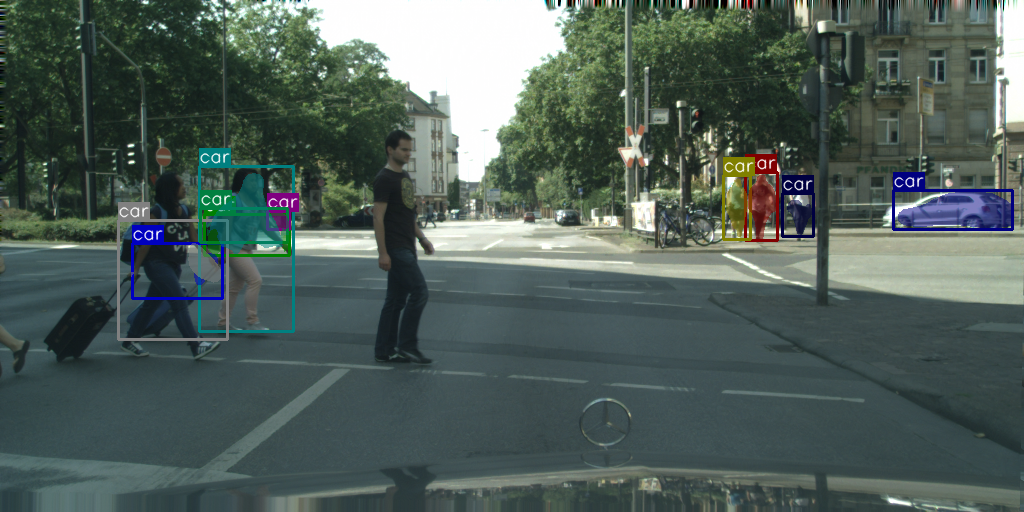}
        \end{subfigure}%
        \hspace*{0.5mm}
        \begin{subfigure}{0.45\textwidth}
          \centering
          \includegraphics[width=1\linewidth]{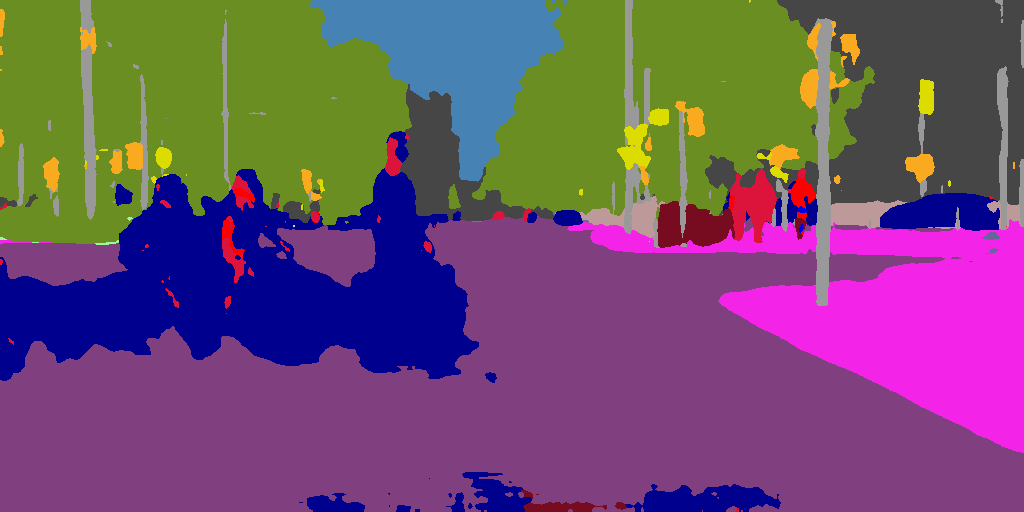}
        \end{subfigure}%
        
        \vspace{2mm}
        \begin{subfigure}{0.45\textwidth}
          \centering
          \includegraphics[width=1\linewidth]{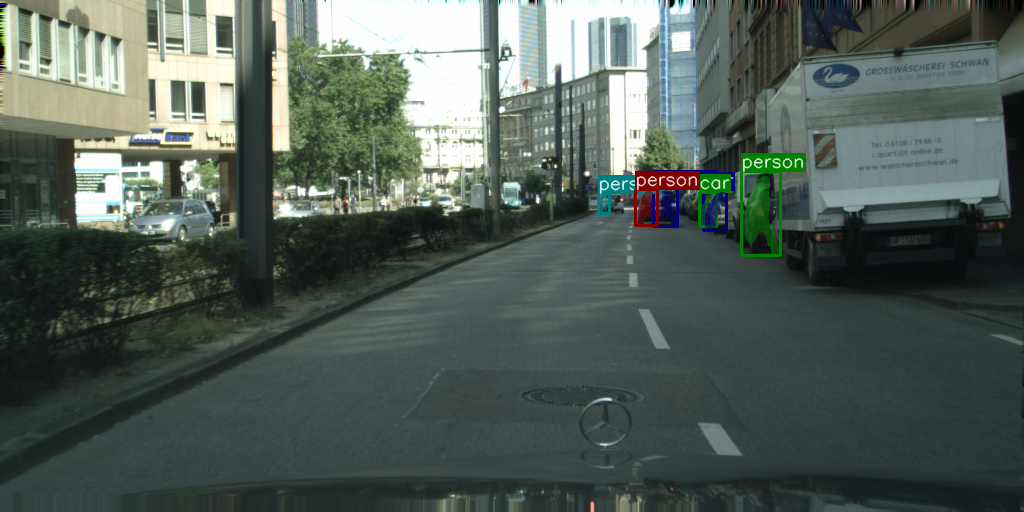}
        \end{subfigure}%
        \hspace*{0.5mm}
        \begin{subfigure}{0.45\textwidth}
          \centering
          \includegraphics[width=1\linewidth]{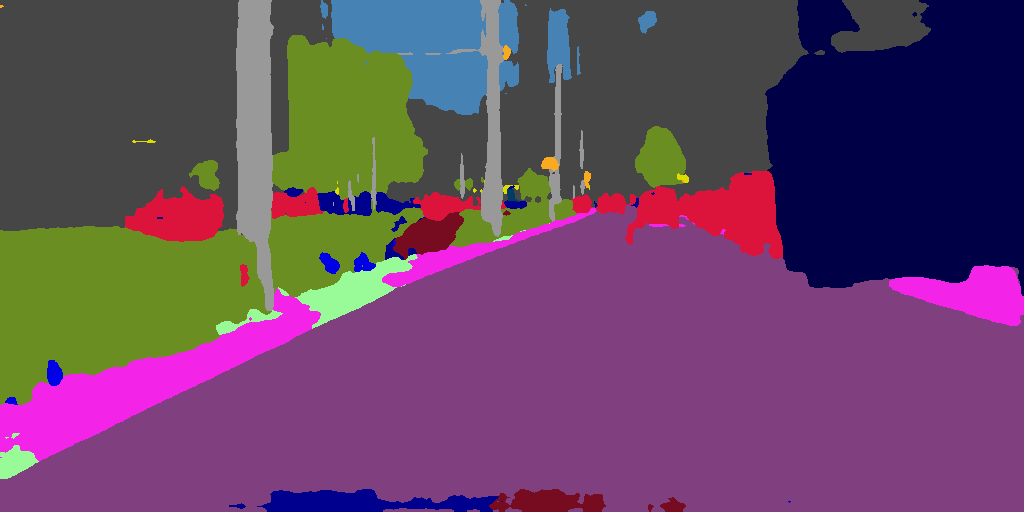}
        \end{subfigure}%
        
  \vspace*{1mm}
  \caption{DAG attack on object detection (first column) and semantic segmentation (second column), respectively. Best viewed in color.}
\label{fig:SM_dagadd}
\end{figure*}

Figure \ref{fig:SM_dagadd} shows the predictions after DAG attack. In object detection predictions (column one), most of the ``car" predictions are horizontal or approximately square noting that DAG attack only targets the classification of boxes. However, there are exceptions such as in row three where two ``persons" on the right of the image are predicted as ``cars" but the boxes are vertical (yellow and red boxes). Corresponding to column-1 images, column-2 shows the segmentation maps when DAG attack is performed on segmentation. We see that the attack manages to switch ``person" and ``car" as ``persons" expected to be red are now blue and ``cars" expected to be blue are now red (the expected color legend is shown in the top of the figure). We see that these predictions also show potential bias towards shape. This bias is particularly evident in row three column-2 where the ``persons" have been combined together in a manner likely resembling the lower portion of a car. However, a systematic approach to investigate this bias is required. 


\section{Ablation study}

The findings presented thus far are based on only the UniNet architecture. To check whether the findings are architecture independent, we choose MTI-Net because it has the best performance in segmentation and depth. We add the instance head to MTI-Net and call this new architecture as MTI-Net++. To be more comparable with UniNet, we did not include the auxiliary tasks used by MTI-Net. The performance of the two architectures on five tasks is shown in Table \ref{table:SM_ablate}.

\begin{figure*}
    \centering
        \begin{subfigure}{.39\textwidth}
          \centering
          \includegraphics[width=1\linewidth]{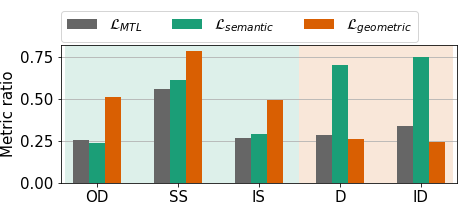}
          \caption{Multi-task, semantic and geometric loss}
          \label{fig:SM_pgdgeoseg_mti}
        \end{subfigure}%
        \begin{subfigure}{.61\textwidth}
          \centering
          \includegraphics[width=1\linewidth]{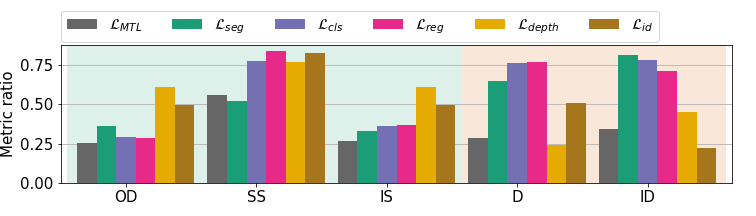}
          \caption{Multi-task and single task loss}
          \label{fig:SM_pgdall_mti}
        \end{subfigure}
   \caption{PGD attacks using different losses with $\epsilon$=1 as $l_\infty$ perturbation bound. Each bar represents a PGD attack conducted on MTI-Net++ with a specific loss (indicated with the bar color). The green shaded and red shaded region contains the metric ratios of semantic tasks and geometric tasks, respectively. Best viewed in color.}
\label{fig:SM_pgd_mti}
\end{figure*}

We repeat the PGD, semantic category hiding and DAG attacks on MTI-Net++. The results of PGD attack with $\epsilon$=1 is provided in the Table \ref{table:SM_pgd5tasks_mti}. Figure \ref{fig:SM_pgd_mti} shows the metric ratios of various tasks after PGD attack with different losses. PGD with semantic loss affects semantic tasks the most and geometric loss affects geometric tasks the most. With individual task losses, the corresponding tasks get affected the most.

\begin{figure*}
    \centering
        \begin{subfigure}{1\textwidth}
          \centering
          \includegraphics[width=1\linewidth]{images/lindau_000025_000019_clean.png}
          \caption{Clean image predictions}
          \label{fig:SM_clean}
        \end{subfigure}%
        
        \begin{subfigure}{1\textwidth}
          \centering
          \includegraphics[width=1\linewidth]{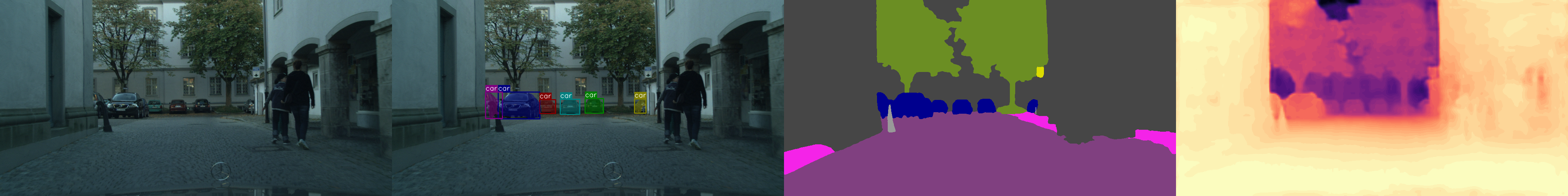}
          \caption{Hide ``persons" from segmentation map}
          \label{fig:SM_segremove}
        \end{subfigure}%
        
        \begin{subfigure}{1\textwidth}
          \centering
          \includegraphics[width=1\linewidth]{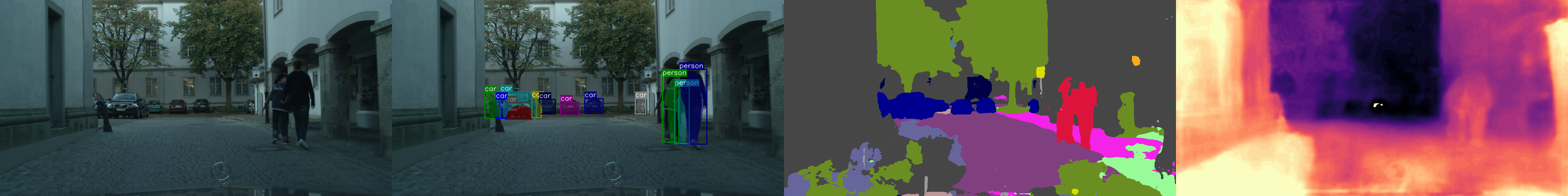}
          \caption{Hide ``persons" from depth map}
          \label{fig:SM_depthremove}
        \end{subfigure}
  \caption{Semantic category hiding with $\epsilon$=2 as $l_\infty$ perturbation bound using MTI-Net++. (a) shows predictions on clean image. Hiding ``persons" from segmentation map in (b) also hides ``persons" from all task predictions. However, in (c) similar effect is not observed when hiding ``persons" from depth map. Starting from the left, the columns show the input image, the instance predictions, the predicted segmentation map and the predicted depth map. Best viewed in color.}
\label{fig:SM_semanticremove_mti}
\end{figure*}

The visualizations of predictions obtained with semantic category hiding attacks (segmentation attack and depth attack) on MTI-Net++, shown in Figure \ref{fig:SM_semanticremove_mti}, resembles that obtained with UniNet. Table \ref{table:SM_semhide_mti} provides the metric values and metric ratios of all tasks after semantic category hiding attacks. Segmentation attack has more effect on depth compared to the effect of depth attack on segmentation.

\begin{figure*}
    \centering
        \begin{subfigure}{.4\textwidth}
          \centering
          \includegraphics[width=0.95\linewidth]{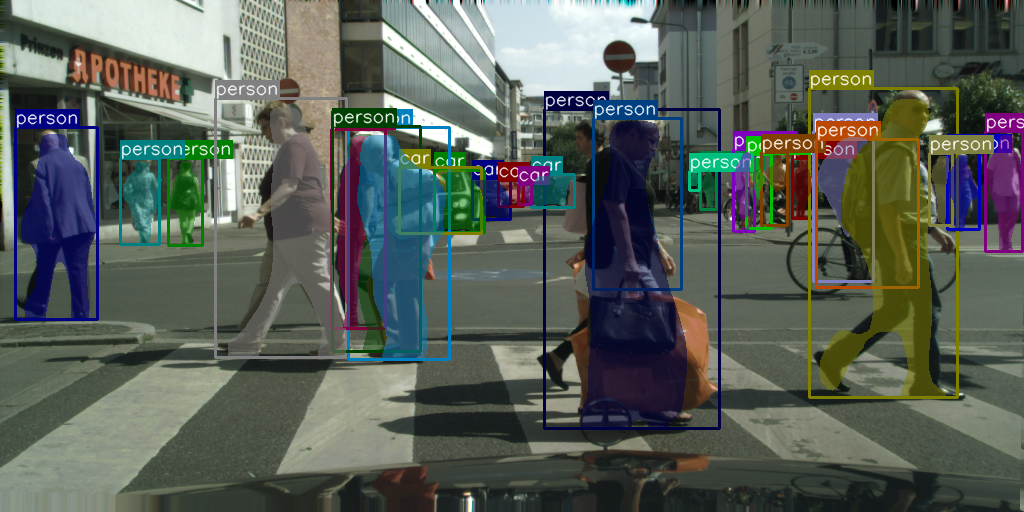}
          \caption{Predictions on clean image.}
          \label{fig:SM_biasorig}
        \end{subfigure}
        \begin{subfigure}{.4\textwidth}
          \centering
          \includegraphics[width=0.95\linewidth]{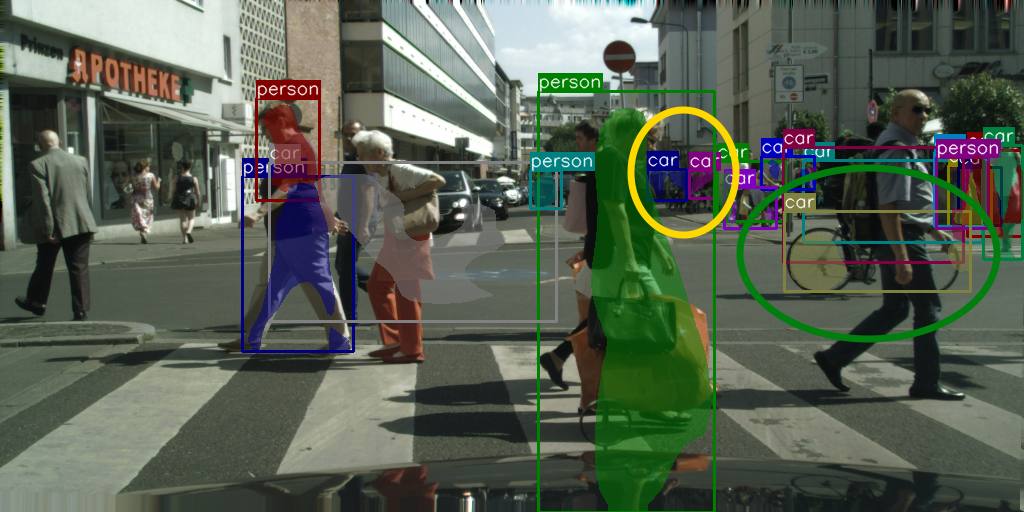}
          \caption{Predictions on DAG generated adversarial image.}
          \label{fig:SM_biasadv}
        \end{subfigure}
  \caption{Shape bias induced by intra-task relationship in object detection. The MTI-Net++ architecture is used. Object detection and instance segmentation predictions visualized on (a) clean image and (b) DAG generated adversarial image. In (b), ``car" box predictions on ``persons" have horizontal aspect ratio despite that the attack objective only includes box classification. The visualization only includes ``car" and ``person" predictions. The rest of the predictions are not visualized. Best viewed in color.}
\label{fig:SM_shapebias_mti}
\end{figure*}

\begin{figure}[t!]
    \centering
    \includegraphics[width=0.85\linewidth]{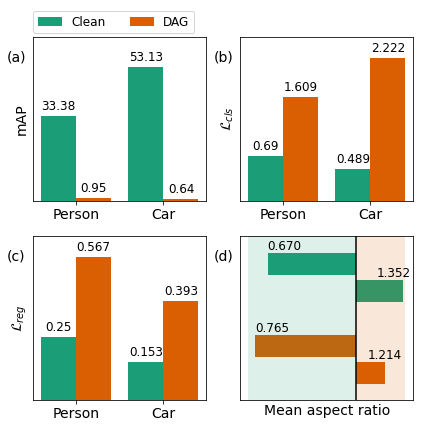}
   \caption{Effect of DAG attack on object detection. Each plot compares before and after DAG attack results specific to the ``person" and ``car" class. Plots (a), (b) and (c) show the mAP, classification loss and regression loss, respectively. Plot (d) shows the mean aspect ratio of ``person" class (green shaded region) and ``car" class (red shaded region).}
\label{fig:SM_shapemetrics_mti}
\end{figure}

Similar to UniNet results, the shape bias learned due the intra-task relationship in object detection is also retained when MTI-Net++ is attacked with the modified DAG attack. Figure \ref{fig:SM_shapebias_mti} shows the visualization of ``person" and ``car" predictions before and after attack. ``person" boxes retain vertical orientation while ``car" boxes retain horizontal or approximately square orientation. Figure \ref{fig:SM_shapemetrics_mti} shows that the aspect ratio of the ``person" and ``car" predictions are preserved even though their class is changed by the DAG attack.

These observation made using MTI-Net++ with all three attacks are in line with the observations made using UniNet, providing strong evidence to the generalizability of findings across architectures.

\begin{table}
\centering
\small
\begin{tabular}{|c|c|c|}
\hline
Tasks & UniNet & MTI-Net++ \\ \hline
\makecell{OD ($\text{mAP}^b$)} & \textbf{38.93}{\scriptsize$\pmb{\pm}$\textbf{0.14}} & 37.12{\scriptsize$\pm$0.20} \\ \hline
\makecell{SS (mIoU)} & 73.85{\scriptsize$\pm0.15$} & \textbf{75.77}{\scriptsize$\pmb{\pm}\textbf{0.40}$} \\ \hline 
\makecell{IS ($\text{mAP}^m$)} & 22.96{\scriptsize$\pm0.09$} & \textbf{24.05}{\scriptsize$\pmb{\pm}\textbf{0.34}$} \\ \hline 
\makecell{D (RMSE)} & 5.52{\scriptsize$\pm0.02$} & \textbf{5.15}{\scriptsize$\pmb{\pm}\textbf{0.03}$} \\ \hline
\makecell{ID ($l_1$ loss)} & \textbf{8.29}{\scriptsize$\pmb{\pm}\textbf{0.07}$} & 9.70{\scriptsize$\pm0.06$} \\ \hline
\end{tabular}
\vspace*{1mm}
\caption{The performace of MTI-Net++ compared with UniNet. MTI-Net++ is the modified version of MTI-Net with the instance head.}
\label{table:SM_ablate}
\end{table}

\begin{table}
	\centering
	\resizebox{\columnwidth}{!}{%
	\begin{tabular}{|c|ccc|cc|}
	\hline
	\multirow{ 2}{*}{\makecell[l]{Loss used \\ for attack}} & OD & SS & IS & D & ID \\ \cline{2-6}
	 & $\text{mAP}^b$ & mIoU & $\text{mAP}^m$ & RMSE & $l_1$ loss \\ \hline 
	 $\mathcal{L}_{MTL}$ & 9.53 & 42.23 & 6.51 & 18.21 & 28.56 \\ \hline
	 $\mathcal{L}_{semantic}$ & 9.02 & 45.96 & 7.10 & 7.36 & 13.08 \\ 
	 $\mathcal{L}_{geometric}$ & 19.01 & 58.89 & 12.12 & 19.51 & 40.14 \\ \hline
	 $\mathcal{L}_{seg}$ & 13.42 & 39.03 & 8.05 & 8.00 & 12.02 \\ 
	 $\mathcal{L}_{cls}$ & 10.97 & 58.40 & 8.81 & 6.80 & 12.56  \\ 
	 $\mathcal{L}_{reg}$ & 10.58 & 62.80 & 8.99 & 6.73 & 13.78 \\ 
	 $\mathcal{L}_{is}$ & 18.60 & 64.03 & 10.74 & 6.43 & 11.84 \\ 
	 $\mathcal{L}_{depth}$ & 22.80 & 57.89 & 14.90 & 21.60 & 21.76 \\ 
	 $\mathcal{L}_{id}$ & 18.59 & 61.92 & 12.13 & 10.18 & 44.31 \\ \hline
     
	\end{tabular}}
	\vspace*{1mm}
	\caption{PGD attack results with $\epsilon$=1 on MTI-Net++ model using different losses.}
	\label{table:SM_pgd5tasks_mti}
\end{table}

\begin{table}
\centering
\small
\begin{tabular}{|c|c|cc|cc|}
\hline
\multirow{ 2}{*}{Tasks} & \multirow{ 2}{*}{Clean} & \multicolumn{ 2}{c|}{Hide from} & \multicolumn{ 2}{c|}{Hide from (rel.)} \\ \cline{3-6}
 &  & \makecell{SS} & \makecell{D} & \makecell{SS} & \makecell{D} \\ \hline
\makecell{OD ($\text{mAP}^b$)} & 33.38 & 6.62 & 13.24 & 0.20 & 0.39 \\ 
\makecell{IS ($\text{mAP}^m$)} & 17.75 & 1.90 & 6.28 & 0.11 & 0.35 \\  
\makecell{SS (mIoU)} & 79.09 & 10.33 & 50.25 & \color{blue}0.13 & \color{red}0.64  \\ \hline 
\makecell{D (RMSE)} & 3.99 & 9.70 & 37.31 & \color{red}0.41 & \color{blue}0.11  \\ 
\makecell{ID (abs rel.)} & 0.15 & 0.33 & 1.12 & 0.45 & 0.13  \\ \hline
\end{tabular}
\vspace*{1mm}
\caption{Results on semantic hiding of ``person" class using MTI-Net++. The first three columns show the actual metric values while the last two columns show metric ratios which is the performance retained by each task after attack with respect to clean performance.}
\label{table:SM_semhide_mti}
\end{table}

\end{document}